\documentclass[10pt]{article} 
\usepackage[preprint]{rlc}

\usepackage{amssymb}            
\usepackage{mathtools}          
\usepackage{mathrsfs}           
\mathtoolsset{showonlyrefs}     
\usepackage{graphicx}           
\usepackage{subcaption}         
\usepackage[space]{grffile}     
\usepackage{url}                
\usepackage{bbm}
\usepackage{booktabs}
\usepackage{multirow}

\usepackage[ruled, vlined]{algorithm2e}

\SetCommentSty{mycommfont}

\newtheorem{assumption}{Assumption}
\newtheorem{theorem}{Theorem}

\newtheorem{lemma}{Lemma}

\newtheorem{definition}{Definition}

\DeclareMathOperator*{\argmax}{arg\,max}

\newcommand{\expt}{\mathbb{E}}
\newcommand{\prob}{\mathbb{P}}

\title{Reinforcement Learning from Human Feedback without Reward Inference: Model-Free Algorithm and Instance-Dependent Analysis}

\author{Qining Zhang \\
        qiningz@umich.edu\\
        University of Michigan, Ann Arbor 
        \And
        Honghao Wei \\
        honghao.wei@wsu.edu \\
        Washington State University
        \AND Lei Ying \\
        leiying@umich.edu \\
        University of Michigan, Ann Arbor
    }

\begin{document}

\maketitle

\begin{abstract}
In this paper, we study reinforcement learning from human feedback (RLHF)  under an episodic Markov decision process with a general trajectory-wise reward model. We developed a model-free RLHF best policy identification algorithm, called $\mathsf{BSAD}$, without explicit reward model inference, which is a critical intermediate step in the contemporary RLHF paradigms for training large language models (LLM). The algorithm identifies the optimal policy directly from human preference information in a backward manner, employing a dueling bandit sub-routine that constantly duels actions to identify the superior one. $\mathsf{BSAD}$ adopts a reward-free exploration and best-arm-identification-like adaptive stopping criteria to equalize the visitation among all states in the same decision step while moving to the previous step as soon as the optimal action is identifiable, leading to a provable, instance-dependent sample complexity $\tilde{\mathcal{O}}(c_{\mathcal{M}}SA^3H^3M\log\frac{1}{\delta})$\footnote{we use $\mathcal{O}(\cdot)$ to hide instance-independent constants and use $\tilde{\mathcal{O}}(\cdot)$ to further hide logarithmic terms except $\log\frac{1}{\delta}$.} which resembles the result in classic RL, where $c_{\mathcal{M}}$ is the instance-dependent constant and $M$ is the batch size. Moreover, $\mathsf{BSAD}$ can be transformed into an explore-then-commit algorithm with logarithmic regret and generalized to discounted MDPs using a frame-based approach. Our results show: (i) sample-complexity-wise, RLHF is not significantly harder than classic RL and (ii) end-to-end RLHF may deliver improved performance by avoiding pitfalls in reward inferring such as overfit and distribution shift.
\end{abstract}

\section{Introduction}
Reinforcement learning (RL), with a wide range of applications in gaming AIs~\citep{knox2008gameAI,macglashan2017gameAI,warnell2018gameAI}, recommendation systems~\citep{YangLiuYing2022Abandonment,zeng2016recommendation,kohli2013recommendation}, autonomous driving~\citep{wei2023driving,schwarting2018driving,kiran2021driving}
, and large language model (LLM) training~\citep{wu2021llm,nakano2021llm,ouyang2022llm,ziegler2019llm,stiennon2020llm}, has achieved tremendous success in the past decade. 
A typical reinforcement learning problem involves an agent and an environment, where at each step, the agent observes the state, takes a certain action, and then receives a reward signal. The state of the environment then transits to another state, and this process continues. 
However, most RL advances remain in the simulator environment where the data acquisition process heavily depends on the crafted reward signal, which limits RL from more realistic applications such as LLM, as defining a universal reward is generally difficult.
In recent years, using human feedback as reward signals to train and fine-tune LLMs has delivered significant empirical successes for AI alignment problems and produced dialog AIs such as the ChatGPT~\citep{ouyang2022llm}. This paradigm where the reward of the state and actions is inferred from real human preferences, instead of being handcrafted, is referred to as \emph{Reinforcement Learning from Human Feedback} (RLHF). A typical RLHF algorithm on LLMs involves three steps: (i) pre-train a network with supervised learning, (ii) infer a reward model from human feedback, in the form of comparisons or rankings among trajectories (responses), and (iii) use classic RL algorithm to fine-tune the pre-trained model. An accurate reward model that aligns with human preferences is the key to the superiority of RLHF.

\textbf{Pitfalls of Reward Inference:} However, most reward models are trained on a maximum likelihood estimator (MLE)~\citep{christiano2017rlpreference,wang2023rlhf,saha2023dueling} under Bradley-Terry model~\citep{bradley1952preference}.
This paradigm exhibits pitfalls: (i) the reward models easily over-fit the dataset which produces in-distribution errors, and (ii) the reward models fail to measure out-of-distribution state-action pairs during fine-tuning. Even though attempts such as pessimistic estimations~\citep{zhu2023rlhf,zhan2023rlhf,zhan2023rlhfoffline} and regularity conditions are made to improve the accuracy and consistency of reward models, it remains a question of whether reward inference is indeed required. Can we develop a model-free RLHF algorithm without reward inference, which has provable instance-dependent sample complexity?

\begin{table}[t]
    \centering
    \renewcommand\arraystretch{1.2}
    \begin{tabular}{cccccc}    
    \toprule
    Setting     & Algorithm & Sample Complexity & Space & Instance & Policy     \\
    \midrule
    \multirow{2}*{RL} & MOCA &  $\mathcal{O}\left(\frac{H^3 SA \log\frac{1}{\delta}}{\Delta_{\min}^2 p^{\pi}_{\max}}\right)$ & model-based & dependent & Opt \\
    ~ &  Q-Learning &  $\tilde{\mathcal{O}}\left(\frac{H^4 SA\log\frac{1}{\delta} }{\varepsilon^2}\right)$ & model-free & independent & $\varepsilon$-Opt \\
    \midrule
    \multirow{3}*{RLHF} & P2R-Q &  $\tilde{\mathcal{O}}\left(\frac{H^4 SA \log\frac{1}{\delta}}{\varepsilon^2}\right)$ & model-free & independent & $\varepsilon$-Opt \\
    ~ & PEPS & $\tilde{\mathcal{O}}\left(\frac{H^2 S^2 A \log\frac{1}{\delta}}{\varepsilon^2} + \frac{S^4 H^3 \log^3\frac{1}{\delta}}{\varepsilon}\right)$ & model-based & independent & $\varepsilon$-Opt \\
    ~ & BSAD(Ours) &  $\mathcal{O}\left(\frac{H^3 M SA^3 \log\frac{1}{\delta}}{(\overline{\Delta}_{\min}^M p^{\pi}_{\max})^2}\right)$ & model-free & dependent & Opt \\
    
    \bottomrule
    \end{tabular}
    \caption{Comparison of RL and RLHF algorithms with MOCA~\citep{wagenmaker2022PACRL}, Q-Learning~\citep{jin2018qlearning}, PEPS~\citep{xu2020preference}, and P2R~\citep{wang2023rlhf} with Q-learning. $S$, $A$, and $H$ are the number of states, actions, and planing steps. $\delta$ is confidence level, $M$ is the batch size. $\Delta_{\min}$ is the minimum value function gap, $\overline{\Delta}_{\min}$ characterizes the preference probability gap (Def.~\ref{def:gap}), and $p^\pi_{\max}$ characterizes the maximum state visitation probability (Def.~\ref{def:visitation}).
    }
    \label{tab:comp}
\end{table}

\textbf{Contributions:} We study an episodic RLHF problem with general trajectory rewards and propose a model-free algorithm called \emph{Batched Sequential Action Dueling} ($\mathsf{BSAD}$) which identifies the optimal action for each state backwardly using action dueling with batched trajectories to obtain human preferences. To equalize the state visitation of the same planning step, we adopt a reward-free exploration strategy and adaptive stopping criteria, which enables learning the exact optimal policy with an instance-dependent sample complexity (Theorem.~\ref{thm:sc_episodic}) similar to classic RL with reward~\citep{wagenmaker2022PACRL}, as long as the batch size is chosen carefully. Moreover, our results only assume the existence of a uniformly optimal stationary policy and do not require the existence of a Condorcet winner, as we will show the optimal policy is the Condorcet winner when human preferences are obtained with large batch sizes. To the best of our knowledge, $\mathsf{BSAD}$ is the first RLHF algorithm with instance-dependent sample complexity, and a transformation of $\mathsf{BSAD}$ will provide the first model-free explore-then-commit RLHF algorithm with logarithmic regret.

\textbf{Comparison to \citep{xu2020preference}:} From the best of our knowledge, the only algorithm with no reward inference (explicit/implicit) is $\mathsf{PEPS}$~\citep{xu2020preference}. Our paper is different in (i) $\mathsf{BSAD}$ is model-free and takes $\mathcal{O}(SA^2)$ space complexity, while $\mathsf{PEPS}$ is model-based and takes $\mathcal{O}(S^2A^2)$ space complexity, (ii) $\mathsf{BSAD}$ employs adaptive stopping criteria which leads to an instance-dependent sample complexity with improved dependence in $S$ and $\delta$, while $\mathsf{PEPS}$ uses fixed exploration horizon and only has worst-case bounds, (iii) we assume the trajectory reward and require the existence of uniformly deterministic optimal policy which slightly generalizes the classic reward, while $\mathsf{PEPS}$ requires the existence of Condorcet winner and stochastic triangle inequality, and (iv) we also generalize to discounted MDPs. The complete comparison of $\mathsf{BSAD}$ and related algorithms is summarized in Tab.~\ref{tab:comp}, and a thorough review of related work is deferred to the appendix.

\section{Preliminaries}

\textbf{Episodic MDP:} An episodic Markov decision process (MDP) is a tuple $\mathcal{M} = (\mathcal{S},\mathcal{A}, H, P, \mu_0)$, where $\mathcal{S}$ is the state space with $|\mathcal{S}| = S$, $\mathcal{A}$ is the action space with $|\mathcal{A}| = A$, $H$ is the planning horizon, $P = \{P_h\}_{h=1}^H$ is the transition kernels, and $\mu_0$ is the initial distribution. At each episode $k$, the agent chooses a policy $\pi^k$, which is a collection of $H$ functions $\{\pi_h^k:\mathcal{S}\to\mathcal{A}\}_{h=1}^H$, and nature samples an initial state $s_1^k$ from the initial distribution $\mu_0$. Then, at step $h$, the agent takes an action $a_h^k = \pi^k_h(s_h^k)$ after observing state $s_h^k$. The environment then moves to a new state $s_{h+1}^k$ sampled from the distribution $P_h(\cdot|s_h^k,a_h^k)$ without revealing any feedback. After each episode, the trajectory of all state-action pairs is collected, which we use $\tau^k$ to denote, i.e., $\tau^k = \tau_{1:H}^k = \{(s_h^k, a_h^k)\}_{h=1}^H$.

\textbf{Trajectory Reward Model:} In this paper,
we assume the expected reward of each trajectory $\tau$ is a general function $f(\tau)$ which maps trajectory to real values, a slight generalization of the cumulative reward structure.
Let $\Psi$ be the set of all partial or complete trajectories.
Then, we assume there exists a function $f:\Psi\rightarrow [0, D]$ which is the expected reward of the MDP $\mathcal{M}$, where $D$ is a positive constant. The reward of a certain trajectory may be random, but humans will evaluate trajectories based on the expected reward. The cumulative reward model is $f(\tau) = \sum_{h=1}^H r(s_h, a_h)$. Under the trajectory reward, we can formulate the Q-function as follows:
\begin{align*}
    V_h^\pi(s) =& \expt^\pi\left[ \left. f(\tau_{h:H})\right| s_h = s\right] = \expt\left[ \left. f(\tau_{h:H})\right| s_h = s, a_h = \pi(s), \tau_{h+1:H} \sim \pi \right],\\
    Q_h^\pi(s,a) =& \expt^\pi\left[ \left. f(\tau_{h:H})\right| s_h = s, a_h = a \right] = \expt\left[ \left. f(\tau_{h:H})\right| s_h = s, a_h = a, \tau_{h+1:H} \sim \pi \right].
\end{align*}
The optimal policy $\pi^*$ is defined as $\pi^*= \argmax_\pi \expt_{\mu_0}[V_1^{\pi}(x_1)]$. 
Without regularity on $f$, learning the $\pi^*$ may fundamentally take $\Omega(A^H)$ samples. Therefore, we impose the following assumption:
\begin{assumption}\label{assump:optimal}
    There exists a uniformly optimal deterministic stationary policy $\pi^*$ for the MDP, i.e.,
    $
        \pi^* = \argmax_\pi V_h^{\pi}(s), \forall (h,s).
    $
\end{assumption}
Under the assumption, we define the value function gap for sub-optimal actions similar to classic MDPs as 
$
    \Delta_h(s,a) = V_h^*(s) - Q_h^*(s,a) = \max_{a'} Q_h^*(s,a') - Q_h^*(s,a).
$
Let $\Delta_{\min} = \min_{h,s,a\neq \pi^*(s)} \Delta_h(s,a)$. For simplicity, we assume the optimal action $\pi^*_h(s)$ is unique for each $(h,s)$. Otherwise, we can incorporate $\Delta_{\min}$ into the algorithm so that the duel between the two optimal actions will terminate in a finite time. As a special case, Convex MDPs~\citep{zahavy2021reward}, e.g., pure exploration~\citep{hazan2019provably}, apprenticeship learning~\citep{abbeel2004apprenticeship}, and adversarial RL~\citep{rosenberg2019adversarial}, satisfy Assumption~\ref{assump:optimal} when the optimal policy is deterministic.

\textbf{Human Feedback:} The agent has access to an oracle (a human expert) that evaluates the average quality (reward) of two trajectory batches. At the end of each episode, the agent has the opportunity to choose two sets of (partial) trajectories, denoted by $\mathcal{D}_0$ and $\mathcal{D}_1$ with cardinality $M_0$ and $M_1$, to query the human for which has the higher average reward. We slightly abuse the notation $\tau$ to let $\tau_0^i$ and $\tau_1^i$ be the $i$-th (partial) trace in $\mathcal{D}_0$ and $\mathcal{D}_1$ respectively, i.e., $\mathcal{D}_0 = \{\tau^1_0, \tau^2_0, \cdots, \tau^{M_0}_0\}$, and $\mathcal{D}_1 = \{\tau^1_1, \tau^2_1, \cdots, \tau^{M_1}_1\}$. Each of them may contain only certain steps. 
After observing the two sets of trajectories, the oracle will give a one-bit feedback $\sigma\in\{0,1\}$ to the agent to indicate the dataset he/she favors. 
For simplicity, let $\overline{f}(\mathcal{D}_1)$ and $\overline{f}(\mathcal{D}_0)$ denote the average trajectory reward of $\mathcal{D}_1$ and $\mathcal{D}_0$. Existing works mostly assume the Bradley-Terry model for preference generalization, i.e., the preference probability is a logistic function of the reward difference, i.e.,
$$
    \prob\left( \mathcal{D}_1 \succ \mathcal{D}_0 \right) = u\left(\overline{f}(\mathcal{D}_1) - \overline{f}(\mathcal{D}_0)\right) = \frac{1}{1+\exp\left( \overline{f}(\mathcal{D}_1) - \overline{f}(\mathcal{D}_0) \right)},
$$
where $u:\mathbb{R}\to [0,1]$ is referred to as the link function~\citep{bengs2021duelingsurvey} which characterizes the structure of preference models. Other link functions, such as linear function, probit function, cloglog function, and cauchit function, have also been well-studied in dueling bandits~\citep{ailon2014duelinglink} and generalized linear models~\citep{razzaghi2013probit,mccullagh2019generalized}, but not RLHF. In this paper, we use a $0$-$1$ link function that indicates the favored set with higher reward, i.e.,
$$
    \sigma = \mathsf{HumanFeedback}(\mathcal{D}_0, \mathcal{D}_1) 
    = \argmax_{i \in \{0,1\}} \overline{f}(\mathcal{D}_i)
    =\argmax_{i \in \{0,1\}} \frac{1}{M_i}\sum_{m=1}^{M_i} f(\tau^m_i) .
$$
Generalization to other link functions can be achieved through revising the probability gap definition below (Def.~\ref{def:gap}). Furthermore, we show in Fig.~\ref{fig:condorcet} that single trajectory preference may not align with the expected reward and thus batched comparison is necessary, and it may be easier for humans to identify a better response if the trajectory batches resemble each other with the same initial state, which motivates the comparison between partial and batched trajectories. Typically, in an LLM training setting, for each candidate policy, the human evaluator will look at multiple responses generated respectively and then assess which policy has a better average quality. Similarly for UAV training, humans will watch multiple UAV trajectories for each policy and declare which policy is better based on the average quality of the movement, i.e., success rate, stability, etc. When the batch sizes are not unbearably large, batched preference assessment of trajectories should not be essentially harder than single trajectory preference assessment.

\textbf{Problem Formulation:} Our goal is to design a learning algorithm to interact with the MDP and learn the optimal policy $\pi^*$ from the human feedback as quickly as possible. 
A learning algorithm $\mathsf{Alg}$ consists of (i) a sampling rule which decides which policy to choose at each episode and whether to query the human agent, (ii) a stopping rule which decides a stopping time when the learner wishes to output an learned policy, and (iii) a decision rule which decides which policy $\hat{\pi}$ to output.
We call an algorithm $\delta$-PAC if it outputs an optimal policy with probability at least $1-\delta$. Our goal is to design such an algorithm to minimize sample complexity $K$:
$$
    \min \expt[K], \text{ such that } \prob(\hat{\pi} = \pi^*) \geq 1-\delta.
$$

\section{Main Results for Episodic MDPs}

In this paper, we focus on the instance-dependent performance. To characterize the structure of the MDPs under human feedback, we introduce the notion of probability gaps in Def.~\ref{def:gap} for each state and sub-optimal action, which is a generalization of the calibrated pairwise preference probability considered in the dueling bandits literature~\citep{yue2012duelingIF,yue2011duelingBTM}. We also define the state visitation probability $p_h^\pi(s)$ of a given policy $\pi$ in Def.~\ref{def:visitation}.

\begin{definition}[Probability Gap]\label{def:gap}
    Given $(h,s)$ and a sub-optimal action $a$, the probability gap $\overline{\Delta}^M_h(s,a)$ for comparison of two trajectory sets with cardinality both being $M$ is defined as:
    \begin{align*}
        \overline{\Delta}^M_h(s,a) = \underbrace{\mathbb{P}\left( \left. \sum_{m=1}^M f(\tau^m_0) > \sum_{m=1}^M f(\tau^m_1) \right| \tau_0^m \sim \pi^*,~ \tau_1^m \sim \{a_h = a, \pi^*\} \right)}_{\overline{p}_h^M(s,a)} - \frac{1}{2},
    \end{align*}
    where the traces $\{\tau^1_0, \cdots, \tau^M_0\}$ are independently sampled starting from state $(h,s)$ using the optimal policy $\{\pi^*_k\}_{k=h}^H$, while $\{\tau^1_1, \cdots, \tau^M_1\}$ are independently sampled starting from state $(h,s)$ using immediate action $a_h = a$ and the optimal policy $\{\pi^*_k\}_{k=h+1}^H$ afterwards. Let $\overline{\Delta}_{\min}^M = \min_{h,s,a}\overline{\Delta}^M_h(s,a)$.
\end{definition}
\begin{definition}[State Visitation Probability]\label{def:visitation}
    Given $(h,s)\in [H]\times \mathcal{S}$, the visitation probability (occupancy measure) of policy $\pi$ is defined as follows:
    \begin{align*}
        p_h^\pi(s) = \prob\left( s_h = s | s_0 \sim \mu_0,~ a_{h'} \sim \pi(s_{h'}),~ \forall h' <h\right).
    \end{align*}
\end{definition}
Let $p^{\pi}_{\max} = \min_{h,s}\max_{\pi}p_h^\pi(s)$, and we assume it is positive. We will use both the probability gap and the state visitation probability to characterize our instance-dependent performance.

\subsection{Algorithm for Episodic RLHF}
\begin{algorithm}[t]
\caption{BASD for Episodic MDPs}\label{alg:BSAD_Episode}
initialize for all $(h,s,a)$, $J_h(s,a)\leftarrow 1$, $L_h(s,a)\leftarrow 0$, $M_{h}(s) \leftarrow 0$, and $l \leftarrow H$, $k \leftarrow 0$ \;
initialize for all $(h,s,a,a')$, $w_{h}(s, a, a') \leftarrow 0$, $N_{h}(s, a, a') \leftarrow 0$, $\hat{\pi}_h(s) = \mathcal{D}^0_h(s) = \mathcal{D}^1_h(s) = \emptyset$ \;
define $\iota \equiv c\log( \frac{SAHk}{\delta})$, $\beta_t = \sqrt{\frac{H\iota}{\max\{t,1\}}}$, and $\alpha_t = \frac{H+1}{H+t}$\;
$\hat{\sigma}_h(s,a,a') \equiv \frac{w_h(s,a,a')}{N_h(s,a,a')}$ or $\frac{1}{2}$ if $N_h(s,a,a') =0$, $b_h(s,a,a') \equiv \sqrt{\frac{\iota}{\max\{N_h(s,a,a'),1\}}} $, $\forall(h,s,a,a')$ \;
    \While{$l\geq 1$}{ 
        receive $s_1$, $k=k+1$\;
        \For(\tcp*[f]{reward-free exploration}){step $h=1:l-1$}{
            take action $a_h\leftarrow \argmax_a J_h(s_h,a)$ and observe $s_{h+1}$, $L_h(s_h, a_h) \leftarrow L_h(s_h, a_h)+1$ \;
            $W_{h+1}(s_{h+1}) \leftarrow \min \{1, \max_a J_{h+1}(s_{h+1},a)\}$ \;
            $J_h(s_h,a_h) \leftarrow (1-\alpha_t) J_h(s_h,a_h) + \alpha_t[W_{h+1}(s_{h+1}) + 2\beta_t]$ where $t = L_h(s_h, a_h)$ \;
        }
        $M_l(s_l) \leftarrow M_l(s_l) +1$. $W_l(s_l) \leftarrow \min\{1, b_{M_l(s_l)}\}$ \;
        call action dueling sub-routine $\mathsf{B\text{-}RUCB}(l,s_l, M_l(s_l))$ \tcp*{action dueling}               
        \If{$\forall s, ~ \exists a, ~\text{such that}~ \forall a', ~ \hat{\sigma}_l(s, a, a') - b_l(s, a, a') \geq 0.5$}{
            $\forall s$, $\hat{\pi}_l(s)\in \{a|\forall a', \hat{\sigma}_l(s, a, a') - b_l(s, a, a') \geq 0.5\}$ \;
            $l\leftarrow l-1$. $J_h(s,a)\leftarrow 1$, $L_h(s,a)\leftarrow 0, \forall (h,s,a)$, $k\leftarrow 0$  \tcp*{backward search}
        }   
    }
    \Return $\hat{\pi}$
\end{algorithm}
\begin{algorithm}[t]\caption{B-RUCB: a batched dueling bandits sub-routine}
\label{alg:BRUCB}
\KwIn{step $h$, state $s$, candidate policy $\hat{\pi}$, past visits $M_h(s)$.}
\eIf{$M_h(s) \pmod {2M} \leq M $}{
    \If(\tcp*[f]{select relative optimal arm}){$M_h(s) \equiv 1 \pmod {M}$}{
        $\mathcal{C}_h(s) = \{a|\forall a': \hat{\sigma}_h(s,a,a') + b_h(s,a,a') \geq 0.5\}$, sample $\hat{a}_s$ uniformly from $\mathcal{C}_h(s)$\; 
        $\mathcal{D}_h^0(s) \leftarrow \emptyset$, $\mathcal{D}_h^1(s) \leftarrow \emptyset$\;
    }
    take action $a_h\leftarrow \hat{a}_s$ and observe $s_{h+1}$, and use policy $\hat{\pi}$ for steps afterwards\;
    $\mathcal{D}_h^0(s) = \mathcal{D}_h^0(s) \cup \{(s_h, a_h) , \cdots, (s_H, a_H)\}$\;
}{
    \If(\tcp*[f]{select combating arm based on UCB}){$M_h(s) \equiv 1 \pmod {M}$}{
        $\tilde{a}_s = \argmax_{a\neq \hat{a}_s}\{\hat{\sigma}_h(s,a,\hat{a}_s) + b_h(s,a,\hat{a}_s)\}$\;
    }
    take action $a_h\leftarrow \tilde{a}_s$ and observe $s_{h+1}$, and use policy $\hat{\pi}$ for steps afterwards \;
    $\mathcal{D}_h^1(s) = \mathcal{D}_h^1(s) \cup \{(s_h, a_h) , \cdots, (s_H, a_H)\}$ \;
}
\If(\tcp*[f]{query human every 2M episodes}){$M_h(s) \equiv 0 \pmod {2M}$}{
    query feedback $\sigma = \mathsf{HumanFeedback}\left(\mathcal{D}^0_h(s), \mathcal{D}^1_h(s) \right)$ \;
    $w_h(s, \tilde{a}_s, \hat{a}_s ) \leftarrow w_h(s, \tilde{a}_s, \hat{a}_s ) + \sigma$, $w_h(s, \hat{a}_s, \tilde{a}_s ) \leftarrow w_h(s, \hat{a}_s, \tilde{a}_s) + 1 - \sigma$ \;
    $N_h(s, \tilde{a}_s, \hat{a}_s) = N_h(s, \tilde{a}_s, \hat{a}_s ) + 1$ \;
}
\Return
\end{algorithm}

In this section, we propose an algorithm called $\mathsf{BASD}$ (Alg.~\ref{alg:BSAD_Episode}) to solve the RLHF for episodic MDPs. The algorithm can be divided into two major modules: (i) an action dueling sub-routine generalizing the $\mathsf{RUCB}$ algorithm from the dueling bandits~\citep{zoghi2014rucb}, and (ii) a reward-free exploration strategy to equalize the visitation probability of each state to minimize the overall sample complexity.

\textbf{Backward Action Dueling:} $\mathsf{BSAD}$ identifies the optimal policy for each state using a backward search. The backbone is to employ a batched version of the $\mathsf{RUCB}$ algorithm~\citep{zoghi2014rucb}, called $\mathsf{B\text{-}RUCB}$ in Alg.~\ref{alg:BRUCB}, which is called in step $l$ and controls the action selection policy from step $l$ to $H$.
Namely, it chooses the action $a_l$ at step $l$ using the $\mathsf{RUCB}$ dueling bandits principle and then uses the candidate optimal policy $\hat{\pi}$ for steps afterward. If the policy $\hat{\pi}$ is indeed the optimal policy $\pi^*$, the average reward from step $l$ to $H$ constitutes an unbiased estimator of $Q_h^*(s_l,a_l)$, which resembles dueling bandits.
Different from classic $\mathsf{RUCB}$, we query human feedback every $2M$ episode with batches and we will show later that it allows the optimal action $\pi^*_h(s)$ to be the action favored by the human oracle (Condorcet winner). Moreover, we adopt a stopping rule for each $(h,s)$ that if there exists one action $a$ whose lower confidence bound of the preference probability estimation $\hat{\sigma}_h(s,a,a')$ is larger than half for all other actions, the optimal action is found. 
Specifically, we use $\mathcal{T}_h(s)$ to denote the stopping rule for state $(h,s)$, i.e.,
$
    \mathcal{T}_h(s) = \{\exists a,~ \forall a', ~ \hat{\sigma}_l(s, a, a') - b_l(s, a, a') \geq 0.5\}.
$
Then, the criteria for $l$ to move from $h$ to $h-1$ is equivalent to $\cap_{s=1}^S \mathcal{T}_h(s)$. 
Running $\mathsf{B\text{-}RUCB}$ with the stopping rule identifies the optimal action $\pi^*_l(s)$ for all states at step $l$ with high probability.

\textbf{Reward-free Exploration:} 
To minimize the sample complexity, it is ideal that every state has a similar visitation probability so that action identification can be performed simultaneously for all the states. Our chosen model-free reward-free exploration between step $1$ to step $l-1$
contributes towards this goal. We slightly adapted the $\mathsf{UCBZero}$ algorithm originally proposed in~\citep{zhang2020rewardfree} in our $\mathsf{BSAD}$ algorithm so that the overall algorithm is model-free. This strategic exploration policy will guarantee that we visit each state on step $l$ proportional to the maximum visitation probability over all possible policy $\pi$ starting from the initial distribution. 

\begin{figure}[t]
    \centering
    \includegraphics[width = 0.8\linewidth]{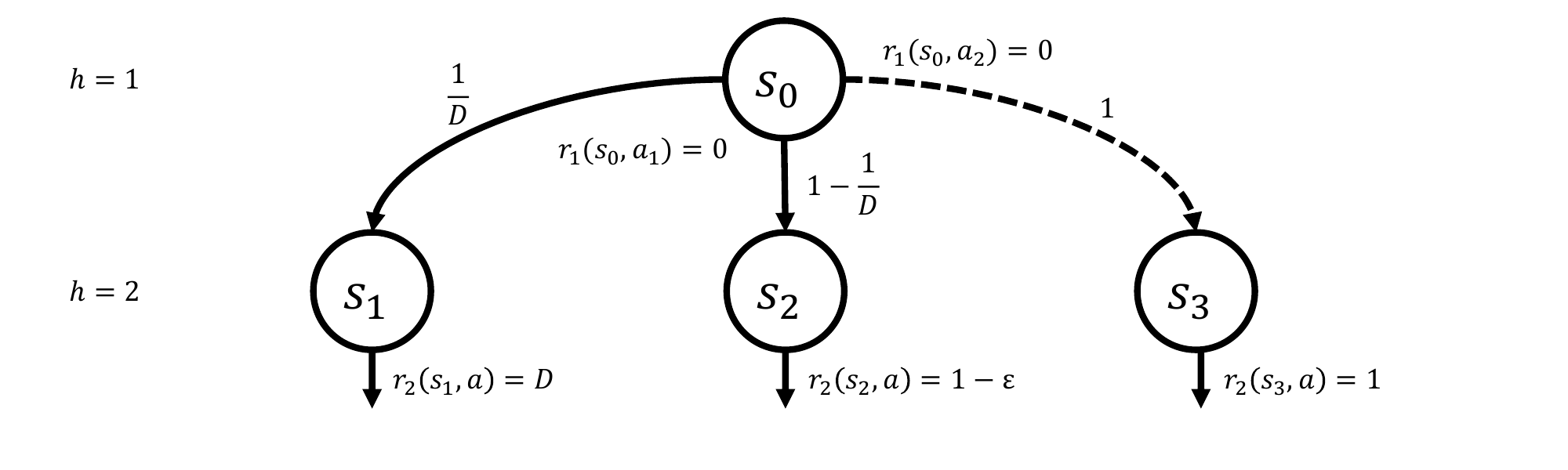}
    \caption{MDP where $\pi^*_h(s)$ is not the Condorcet winner: there are $3$ states ($\{s_1, s_2, s_3\}$) at step $2$ with $1$ action, and $1$ state $s_0$ in step $1$ with $2$ actions. With action $a_1$, the state transits w.p. $1/D$ to state $s_1$ with reward $D$, and w.p. $1-1/D$ to state $s_2$ which gives reward $1-\varepsilon,$ where $0 < \epsilon < 1.$ With action $2$, the state transits deterministically to state $s_3$ with reward $1$.}
    \label{fig:condorcet}
\end{figure}

\subsection{Theoretical Results}
It is well-known from dueling bandits literature~\citep{zoghi2014rucb} that the $\mathsf{RUCB}$ algorithm only requires the existence of the Condorcet winner to identify the optimal action, where the Condorcet winner refers to an action that is preferred with probability larger than half when compared to any other action. Similar to the definition in dueling bandits, for any state $(h,s)$ and any size $M$, we say the optimal action $\pi^*_h(s)$ is the Condorcet winner if the preference probability $\overline{p}_h^M(s,a)$ is larger than half for all other actions $a$. For any comparison-based algorithm to identify the optimal policy, the optimal policy must be the Condorcet winner. We will first characterize the existence of the Condorcet winner when human experts are queried with batch size $M$ large enough.
\begin{lemma}\label{lemma:condorcet}
    Given an MDP $\mathcal{M}$ and for any $(h,s)$, the action $\pi^*_h(s)$ associated with the optimal policy $\pi^*$ is the Condorcet winner in the $\mathsf{HumanFeedback}$ comparison as long as
    $
        M = \Omega(D^2\Delta_{\min}^{-2}).
    $
\end{lemma}

\textbf{Existence of Condorcet Winner:} In general, the optimal action $\pi^*_h(s)$, although it maximizes the expected reward, is not necessarily the Condorcet winner with arbitrary $M$. 
To see this, consider a two-step MDP with traditional cumulative reward as shown in Fig.~\ref{fig:condorcet}. For state $s_0$ and $D>2$ in step $1$, the optimal action is $a_1$ which gives expected reward $1+(1-D^{-1})(1-\varepsilon)$ larger than $1$ given by action $a_2$. However, if we choose $M=1$ and query human feedback of the duel between action $a_1$ and $a_2$, the human expert will only prefer action $a_1$ if the state transits to $s_1$, which only occurs with probability $1/D$ and could be much less than half. Therefore, the optimal action $a_1$ for state $s_0$ is not the Condorcet winner. Similarly, it is also not hard to construct counter-examples with more than three actions to show that the Condorcet winner does not exist. 
However, Lemma.~\ref{lemma:condorcet} shows that the optimal action $\pi^*_h(s)$ is indeed the Condorcet winner at every state $(h,s)$ as long as the batch size $M$ is large enough. The bound is proportional to $D^2$ which characterizes the variance of reward for a single trajectory and inversely proportional to the square of the minimum value function gap $\Delta_{\min}$, which characterizes the distinguishability among actions. 
The proof of Lemma.~\ref{lemma:condorcet} is deferred to the appendix, where we apply concentration inequalities to lower bound the preference probability $\overline{p}_h^M(s,a)$. 
Next, we characterize the sample complexity of $\mathsf{BSAD}$.

\begin{theorem}\label{thm:sc_episodic}
    Given an MDP $\mathcal{M}$, fix $\delta$ and suppose $M$ is chosen large enough such that the optimal policy $\pi^*$ is the Condorcet winner for all states $(h,s)$. Then with probability at least $1-\mathcal{O}(\delta)$, the $\mathsf{BSAD}$ algorithm terminates within $K$ episodes and returns the optimal policy $\hat{\pi}=\pi^*$ with:
    \begin{align*}
        K = \tilde{\mathcal{O}}\left( \sum_{h=1}^H  \frac{SA^3h^2M \log(\frac{1}{\delta})}{\min_{s,a} \max_\pi [\overline{\Delta}_h^M(s,a) p^\pi_h(s)]^2} \right).
    \end{align*}
\end{theorem}
\textbf{Proof Roadmap:} Our main Theorem.~\ref{thm:sc_episodic} conveys two messages: (i) $\mathsf{BSAD}$ is $\delta$-PAC, and (ii) $\mathsf{BSAD}$ has provable instance-dependent sample complexity bound under general reward model. 
The proof of Theorem.~\ref{thm:sc_episodic} is deferred to appendix. To obtain the correctness guarantee, we decompose the probability of making a mistake into the sum of probabilities where the mistake is made on a certain step $h$. Then, using a backward induction argument, we show that the total mistake probability is small. To obtain the sample complexity bound, we fix $(h,s)$ and then bound the number of comparisons between two actions. 
Next, we bound the total number of comparisons and the total number of episodes needed to identify the optimal action for this $(h,s)$. This can be achieved by summing up the number of comparisons between all pairs of arms before the stopping criteria $\mathcal{T}_h(s)$ for that state is satisfied. Lemma.~\ref{lemma:sc_episodic_state} characterizes the sample complexity for any state $(h,s)$:
\begin{lemma}\label{lemma:sc_episodic_state}
    Given an MDP $\mathcal{M}$, fix $\delta$ and suppose $M$ is large enough. For fixed $(h,s)$, the number of episodes with $l=h$ and $s_h = s$ until the criteria $\mathcal{T}_h(s)$ is bounded with high probability by:
    \begin{align*}
        M_h(s) 
        =& \tilde{\mathcal{O}}\left( \sum_{i=2}^A \frac{i}{\overline{\Delta}_h^M(s,a_i)^2} M\log\left(\frac{1}{\delta} \right) \right)
        = \tilde{\mathcal{O}}\left( \frac{A^2 M \log\left(\frac{1}{\delta} \right)  }{\min_{a}\overline{\Delta}_h^M(s,a)^2} \right),
    \end{align*}
    where $\{a_1, a_2, \cdots, a_A\}$ is a permutation of the action set $\mathcal{A}$ such that $a_1$ is the optimal action and $\overline{\Delta}_h^M(s,a_2)\leq \overline{\Delta}_h^M(s,a_2)\leq \cdots, \overline{\Delta}_h^M(s,a_A)$.
\end{lemma}
Notice that our bound in Lemma.~\ref{lemma:sc_episodic_state} is different from the original $\mathsf{RUCB}$ algorithm provided in~\citep[Theorem 4]{zoghi2014rucb} due to (i) we study a PAC setting while the vanilla $\mathsf{RUCB}$ focuses on regret minimization and (ii) we chose a larger confidence bonus so that our bound only have logarithmic dependence on $\delta$. After bounding the sample complexity to identify the optimal action for each state, we need to relate $M_h(s)$ to the total number of episodes through reward-free exploration. 
We show in Lemma.~\ref{lemma:rewardfree_episodic} that the number of episodes spent for a step $l=h$ is bounded by the number of visitations $M_h(s)$, which is analog to~\citep[Theorem 3]{zhang2020rewardfree}. 
\begin{lemma}\label{lemma:rewardfree_episodic}
    Given an MDP $\mathcal{M}$, fix $\delta$ and suppose $M$ is large enough. For a fixed $(h,s)$, suppose we have $l=h$ and $k=K_h$ in the current episode, we have:
    \begin{align*}
        \forall s,~ K_h \leq \mathcal{O}\left(\frac{SAh^2 M_h(s)}{\max_\pi p^\pi_h(s)^2}\right).
    \end{align*}
\end{lemma}
Combining both Lemma.~\ref{lemma:sc_episodic_state} and Lemma.~\ref{lemma:rewardfree_episodic}, we will be able to prove Theorem.~\ref{thm:sc_episodic}:
\begin{align*}
    K = \sum_{h=1}^H \max_s\mathcal{O}\left(\frac{SAh^2 M_h(s)}{\max_\pi p^\pi_h(s)^2}\right) 
    = \tilde{\mathcal{O}}\left( \sum_{h=1}^H  \frac{SA^3h^2M \log(\frac{1}{\delta})}{\min_{s,a} \max_\pi [\overline{\Delta}_h^M(s,a) p^\pi_h(s)]^2} \right).
\end{align*}

\textbf{RLHF Algorithm with Logarithm Regret:} It is very simple to adapt the $\mathsf{BSAD}$ algorithm to an explore-then-commit type algorithm for regret minimization by choosing $\delta = T^{-1}$. Then, the sample complexity bound will convert into a regret bound in the order of $\mathcal{O}(\log T)$. To the best of our knowledge, this is the first RLHF algorithm with logarithmic regret performance.

\textbf{Instance Dependence and Connection to Classical RL:} Our sample complexity bound in Theorem.~\ref{thm:sc_episodic} has a linear dependence on the number of states $S$, a polynomial on the number of actions $A$ and the planning horizon $H$, and a logarithmic dependence on the inverse of confidence $\delta$. Moreover, it characterizes how the sample complexity depends on fine-grained structures of the MDP $\mathcal{M}$ itself. It is also inversely proportional to the square of the probability gap $\overline{\Delta}_h^M(s,a)$ which resembles the sample complexity or regret bounds in the dueling bandit literature, and also resembles the dependence of the value function gap $\Delta_h(s,a)$ in the sample complexity bounds for traditional tabular RL, e.g.,~\citep[Theorem 2]{wagenmaker2022PACRL}. Moreover, the inverse proportional dependence of the maximum state visitation probability over all policies also resembles the traditional RL. In fact, with $M$ chosen in the same order as in Lemma.~\ref{lemma:condorcet} and using concentration inequalities, the sample complexity bound can be converted depending on the value function gap as follows:
\begin{align*}
    K = \tilde{\mathcal{O}}\left( \frac{SA^3H^3D^2 \log(\frac{1}{\delta})}{\min_{h,s,a}\Delta_h(s,a)^2 \max_\pi p^\pi_h(s)^2} \right).
\end{align*}
This shows that RLHF is almost no harder than classic RL given the appropriate parameter, except for a polynomial factor on the number of actions $A$ and the planning horizon $H$. This finding coincides with~\citep{wang2023rlhf} and sheds light on the similarity between RLHF and classic RL. Notice that our result is derived from a general reward model where the Bellman equations do not hold. Therefore, our result also seemingly implies that the fundamental backbone of RL is the existence of uniformly optimal stationary policy instead of the Bellman equations.

\section{Generalization to Discounted MDPs}
In this section, we generalize the $\mathsf{BSAD}$ algorithm to discounted MDPs with the traditional state-action reward function $r(s,a)\in[0,1]$ and discount factor $\gamma$. Our approach is to segment the time horizon into frames with length $H = \Theta( \frac{1}{1-\gamma}\log \frac{1}{\varepsilon (1-\gamma)^2} )$. Then, we run $\mathsf{BSAD}$ (Alg.~\ref{alg:BSAD_Episode}) with horizon $H$ on the discounted MDP, as if it is episodic. This frame-based adaptation delivers provable instance-dependent sample complexity shown in Theorem.~\ref{thm:discounted-main}. Discussions are deferred to the appendix.

\begin{theorem}\label{thm:discounted-main}
    suppose $M$ is chosen large enough. Then with probability $1-\mathcal{O}(\delta)$, $\mathsf{BSAD}$ terminates within $K$ episodes and returns an $\varepsilon$-optimal policy with:
    \begin{align*}
        K = \tilde{\mathcal{O}}\left( \frac{SA^3M \log(\frac{1}{\delta})\log^3(\frac{1}{\varepsilon})}{(1-\gamma)^3\min_{h,s,a} \overline{\Delta}_h^M(s,a) ^2 \max_\pi \min_{s'} p^\pi_h(s|s')^2} \right),
    \end{align*}
    where $\overline{\Delta}_h^M(s,a)$ to be the probability gap for action $a$ and trajectories of length $H-h$ compared to the Condorcet winner of that state $s$, and $p^\pi_h(s|s')$ is the visitation probability of $s$ after $h$ steps starting from state $s'$ with policy $\pi$. Both definitions are analog to the definitions in episodic MDPs.
\end{theorem}

\section{Numerical Results}

In this section, we study the empirical performance of $\mathsf{BSAD}$ on an MDP based on Fig.~\ref{fig:condorcet} with $D=10$. The only difference is we replicate two copies of $s_0$ in the first step with different initial distributions. For these states, the optimal policy is not the Condorcet winner under a single trajectory comparison but will become the Condorcet winner when the batch size increases. We compare  $\mathsf{BSAD}$ to existing value-based model-free RLHF algorithms, with and without reward inference, where the performance is measured by the value function $\expt_{\mu_0}[V_1^{\hat{\pi}}(s)]$ of the candidate policy evaluated on the true MDP. The baselines that we chose are (i) a model-free and batched adaptation of $\mathsf{PEPS}$~\citep{xu2020preference} (no reward inference) which uses $\mathsf{UCBZero}$~\citep{zhang2020rewardfree}, (ii) Q-learning with P2R~\citep{wang2023rlhf} (reward inference) where the candidate policy $\hat{\pi}$ is the greedy policy, and (iii) $\mathsf{REGIME}$~\citep{zhan2023rlhf} (reward inference) with $\mathsf{UCBZero}$ and pessimistic Q-learning~\citep{shi2022pessimistic} as offline RL oracle, where each point is obtained through a $1k$-episode offline RL algorithm. We also compare to classic RL algorithms, i.e., Q-learning~\citep{jin2018qlearning}. 

\begin{figure}[tbp]
    \centering
    \begin{subfigure}[b]{0.32\textwidth}
        \centering
        \includegraphics[width=\textwidth]{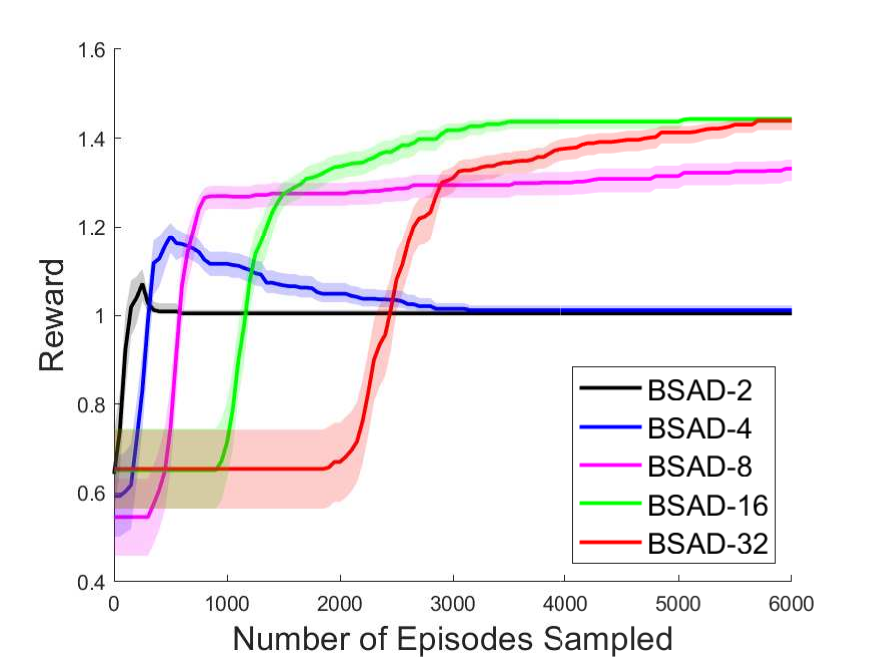}
        \caption{batch size}
        \label{fig:sub1}
    \end{subfigure}
    \begin{subfigure}[b]{0.32\textwidth}
        \centering
        \includegraphics[width=\textwidth]{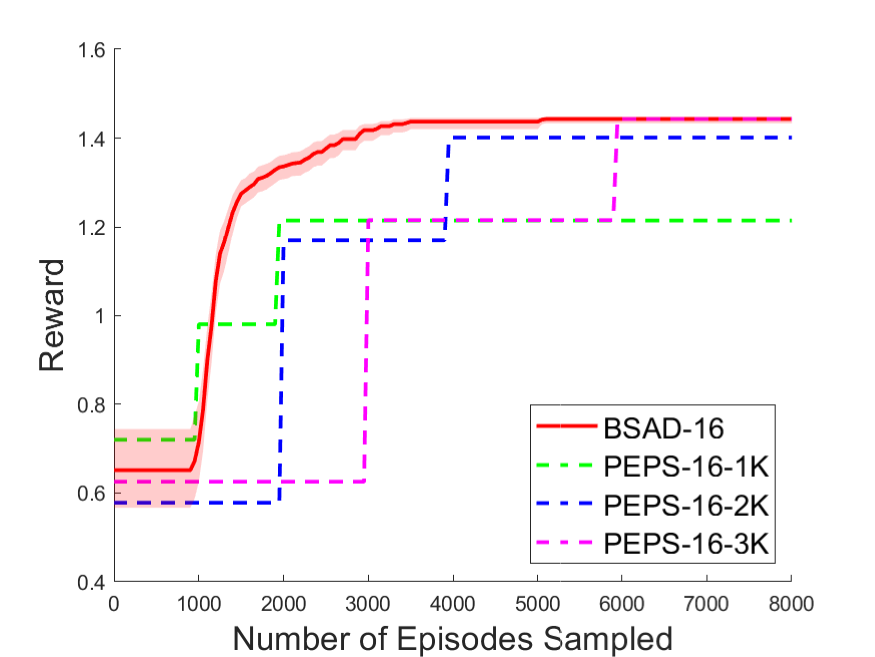}
        \caption{adaptive stopping}
        \label{fig:sub2}
    \end{subfigure}
    \begin{subfigure}[b]{0.32\textwidth}
        \centering
        \includegraphics[width=\textwidth]{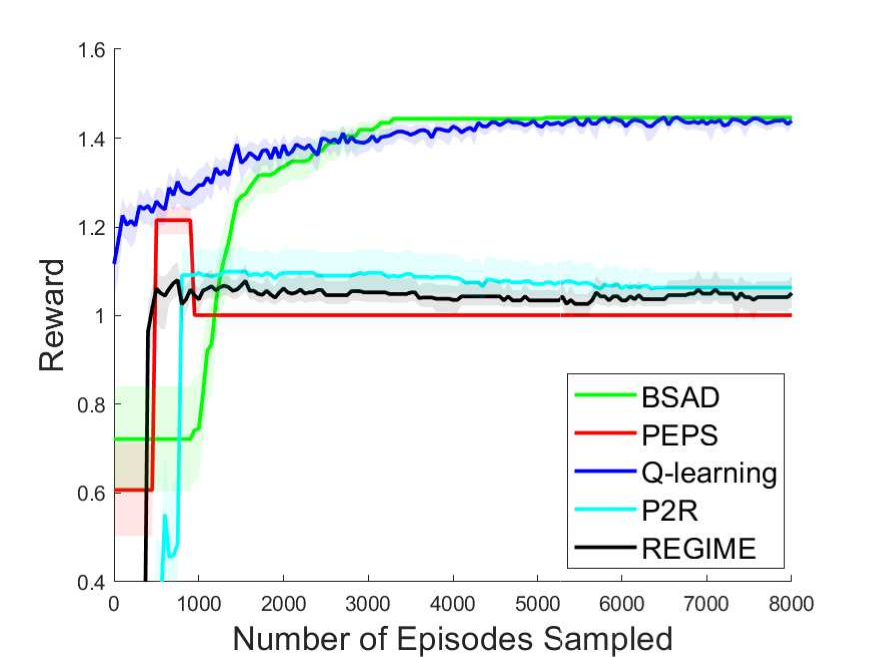}
        \caption{performance}
        \label{fig:sub3}
    \end{subfigure}
    \caption{numerical experiment on a three-state two-step MDP: (a) shows the proposed $\mathsf{BSAD}$ algorithm with different batch sizes. (b) compares $\mathsf{BSAD}$ with adaptive stopping to batched version of $\mathsf{PEPS}$ with fixed exploration horizon. (c) compares $\mathsf{BSAD}$ to model-free RLHF and RL algorithms. Results are averaged over $100$ trajectories and shaded areas represent bootstrap confidence intervals.}
    \label{fig:numerical}
\end{figure}

\begin{table}[htbp]
    \centering
    \begin{tabular}{cccccc}
    \toprule
        Algorithm & BSAD(ours) & PEPS & Q-learning & P2R & REGIME \\
    \midrule
        Running Time (ms) & 171.21 & 179.23 & 1090.12 & 5898.30 & 4613.73 \\
    \bottomrule
    \end{tabular}
    \caption{running time comparisons on $1$ CPU averaged over $50$ trajectories.}
    \label{tab:runningtime}
\end{table}

Fig.~\ref{fig:sub1} shows the effect of batch size. When using a small batch size, i.e., $M=2,4$, the Condorcet winner at $h=1$ is not optimal, and $\mathsf{BSAD}$ converges to a sub-optimal policy. When $M$ is large, $\mathsf{BSAD}$ identifies the optimal policy, and the sample complexity displays a decrease-then-increase trend, which coincides with Theorem.~\ref{thm:sc_episodic}. Specifically, when $M$ increases, the probability gap in the denominator increases sharply, leading to reduced sample complexity, and as $M$ continues to increase, $M$ in the numerator starts to dominate. This justifies $\mathsf{BSAD}$ is adaptive to MDP instances. Fig.~\ref{fig:sub2} shows the comparison of $\mathsf{BSAD}$ to a batched version of $\mathsf{PEPS}$ with different exploration horizons. The observation that the curve of $\mathsf{BSAD}$ lies uniformly above all $\mathsf{PEPS}$ curves shows the necessity of adaptive algorithm design. Specifically, our design of adaptive stopping criteria identifies the optimal policy earlier and adapts to the different distinguishability in different states, which results in improved regret performance. In Fig.~\ref{fig:sub3}, we compare $\mathsf{BSAD}$ to Q-learning and RLHF algorithms with reward inference. First, we observe that $\mathsf{BSAD}$ has almost the same performance as Q-learning which uses the reward information, which shows RLHF is almost no harder than classic RL. However, our algorithm applies to the general trajectory reward function while Q-learning cannot be used anymore. $\mathsf{BSAD}$ exhibits superior performance than other RLHF algorithms also in running time as shown in Table.~\ref{tab:runningtime}, because training reward models with MLE is difficult and takes much larger sample and computational complexity, let alone the best policy can only be obtained when the reward model is accurate enough. This observation somewhat justifies the reward model is unnecessary given it suffers from pitfalls like over-fitting and distribution shift.

\section{Conclusion}
We studied RLHF under both episodic MDPs with trajectory reward structure, a generalization of the classic cumulative reward. We propose an algorithm called $\mathsf{BSAD}$ which enjoys a provable instance-dependent sample complexity that resembles the result in classic RL with reward. We also generalize our results to discounted MDPs. Our results show RLHF is almost no harder than classic RL, and the current dominating reward model training module in RLHF may be unnecessary. 

\subsubsection*{Acknowledgments}\label{sec:ack}
The work of Qining Zhang and Lei Ying is supported in part by NSF under grants 2112471, 2134081, 2207548, 2240981, and 2331780.

\bibliography{ref}
\bibliographystyle{rlc}
\clearpage

\appendix

\section{Related Work}
Reinforcement learning, with a wide range of applications in gaming AIs~\citep{knox2008gameAI,macglashan2017gameAI,warnell2018gameAI}
, recommendation systems~\citep{YangLiuYing2022Abandonment,zeng2016recommendation,kohli2013recommendation}
, autonomous driving~\citep{wei2023driving,schwarting2018driving,kiran2021driving}
, and large language model (LLM) training~\citep{ziegler2019llm},
has achieved tremendous success in the past decade. 
A typical reinforcement learning problem involves an agent and an environment, where at each step, the agent observes the state, takes a certain action, and then receives a reward signal. The state of the environment then transits to another state, and this process continues. In this section, we review works on RL that is relevant to our paper.

\textbf{Best Policy Identification in RL:} best arm identification (BAI) and best policy identification have been studied in both bandits and reinforcement learning contexts for years~\citep{bechhofer1958sequential}. Most work focuses on the fixed confidence setting where one intends to identify the optimal or $\varepsilon$-optimal action as quickly as possible satisfying the probability of making a mistake that is smaller than some constant $\delta$. In \citep{kaufmann2016complexity,Garovoer16BAI}, the authors introduce a non-asymptotic lower bound for BAI. Subsequently, they propose the Track and Stop algorithm ($\mathsf{TAS}$) that achieves this lower bound asymptotically. The $\mathsf{TAS}$ algorithm has since been extended to a variety of other settings \citep{pmlr-v201-jourdan23a, garivier2021nonasymptotic}. Similarly, best policy identification has also been studied in MDPs where people also call this PAC-RL~\citep{dann2015pacrl,dann2017unifying}. However, different from BAI problem, researchers are often satisfied with an $\varepsilon$-optimal policy in PAC-RL, that is a policy which has value function $\varepsilon$ close to the optimal value function. Worst-case performance of PAC-RL can easily be obtained through random policy selection over a low-regret RL algorithm~\citep{jin2018qlearning}, this approach somehow does not require the design of adaptive stopping time and the performance bound depends on $\varepsilon$, which usually does not adapt to the instances. In~\citep{tirinzoni2022pacrl, tirinzoni2023optimistic}, researchers attempted optimal instance-dependent PAC-RL with exact optimal policy identification. They formulated the MDP to a minimax problem similar to BAI literature and proposed an algorithm to almost match the lower bound. However, their instance-dependent constant does not have a closed form which makes the dependence of sample complexity on the MDP structure elusive. Authors of~\citep{wagenmaker2022PACRL} provided an algorithm called $\mathsf{MOCA}$ which uses reward-free exploration with an action elimination algorithm to achieve an instance-dependent sample complexity bound with closed form.

\textbf{Instance-Dependent Analysis in RL:} beyond best policy identification results~\citep{tirinzoni2022pacrl,tirinzoni2023optimistic,wagenmaker2022PACRL}, instance dependent regret bounds are also studied in~\citep{foster2020instance,dann2021beyond,xu2021fine,simchowitz2019instance,yang2021q}, where~\cite{xu2021fine,simchowitz2019instance} focus on model-based algorithms, while \cite{yang2021q} focus on model-free algorithms.

\textbf{Adaptive Stopping Design in RL:} in best arm identification, usually, the key to delivering a promising and close to optimal sample complexity performance is to design the stopping criteria which stops the sampling rule as soon as the best action is identifiable. Before ``model-based'' methods such as TAS~\citep{Garovoer16BAI} which uses complicated Chernoff statistics to control the stopping time, there were many ``confidence-based'' algorithms \citep{2012kl, Kaufmann13kl, UGapE} focused on constructing high-probability confidence intervals. The stopping time is then when one confidence interval is disjoint from and greater than all the rest, which inspires our design. A similar designing principle is also adopted for bandit algorithms with multiple objectives, e.g., regret minimization while identifying the optimal action~\citep{degenne2019ABtest,zhang2024ROBAI}. Within this approach, the algorithms can identify the exact optimal policy, while also adapts the sample complexity to the structure of the problem, involving constants such as the value function gap in the theoretical performance bounds.

\textbf{Dueling Bandits:} The dueling bandit problem is proposed in~\citep{yue2012duelingIF} which studies the problem of identifying the optimal action, or achieving a low-regret performance when only comparison feedback is given. This problem is also called preference-based reinforcement learning, and the reader is encouraged to refer to~\citep{bengs2021duelingsurvey} for a complete survey on this subject. Usually, various comparison or preference assumptions are required so that the dueling bandit algorithm can deliver good performance, for example, the existence of Condorcet winner and stochastic triangle inequality is required in Beat The Mean~\citep{yue2011duelingBTM}, while RUCB~\citep{zoghi2014rucb} only requires the existence of the Condorcet winner. Contextual dueling bandits are also studied in~\citep{dudik2015contextual,saha2022efficient}

\textbf{Reinforcement Learning from Human Feedback:} In recent years, the development of LLMs~\citep{wu2021llm,nakano2021llm,ouyang2022llm,ziegler2019llm,stiennon2020llm} has motivated the study of RLHF~\citep{christiano2017rlpreference,kaufmann2023survey}, which is a generalization of dueling bandits in the MDP setting.~\cite{xu2020preference} studies the RLHF problem in standard tabular MDPs by reducing the RL problem into bandits of each state.~\cite{novoseller2020dueling} generalizes posterior sampling to preference-based RL and~\cite{saha2023dueling} studies RLHF under trajectory feature and preferences. For MDPs with function approximation, most existing work~\citep{saha2023dueling,zhan2023rlhf,zhan2023rlhfoffline,chen2022human,kong2022provably} are based on the Bradley-Terry-Luce model and MLE to characterize the human preferences and transform it into rewards. Specifically, in~\citep{wang2023rlhf}, the authors build a preference to reward interface that uses confidence balls to decide whether to query human feedback.~\cite{zhu2023rlhf} generalizes reward inference from pair-wise comparison to multiple preference rankings.~\citep{kausik2024framework} studies RLHF with partially observed rewards and states. All aforementioned RLHF algorithms are value-based, and~\cite{du2024exploration} studies policy gradient with human feedback. To the best of our knowledge, almost all works in literature adopt the pipeline that infers the reward model and then performs classic RL algorithms. Attempts to bypass reward modeling have also been studied. In~\citep{rafailov2024dpo}, the authors proposed the DPO method which optimizes the policy directly from the preference over trajectories. However, DPO is purely empirical and still assumes the Bradley-Terry model to implicitly infer the reward. Moreover, DPO requires the knowledge of a reference policy and thus is only suitable in the fine-tuning phase. Attempts to understand the theoretical performance of DPO and its generalizations are made in~\citep{azar2024general}, but the authors showed the existence of optima of the loss function, which does not provide any policy optimality and sample complexity guarantees. 

\textbf{Reinforcement Learning with General Feedback:} RL with trajectory-wise feedback is first studied in~\citep{efroni2021trajfeedback} where the authors assume the instant reward of a state-action pair is not observable while the cumulative reward of a trajectory is presented at the end of each episode. ~\cite{chatterji2021theory} studies RL with one-bit "good" or "bad" instructional trajectory feedback, and ~\cite{saha2023dueling} studies dueling RL with trajectory feedback. These works either assume the trajectory feedback is linear through known feature vectors of trajectory, or assume the feedback credit can be assigned to the visited state-action pairs. Another line of work that generalizes the classic MDP assumption is the convex MDPs~\citep{hazan2019provably,zahavy2021reward,mutti2023convex}, where the value function is assumed to be a convex function of the state occupancy measures, instead of a linear function in classic MDPs. It is worth remarking that our Assumption.~\ref{assump:optimal} intersects with the convex MDP assumption, since there exists a uniformly stationary random optimal policy for convex MDPs, but not necessarily deterministic. Besides convex MDPs, ~\cite{prajapat2023submodular} studied submodular RL where the value functions are submodular but not necessarily additive.

\section{Proofs of Results for $\mathsf{BSAD}$ in Episodic MDPs}
In this section, we provide detailed proofs of theoretical results for $\mathsf{BSAD}$ in episodic MDPs. Before we proceed, we first provide several definitions that will be useful in our proofs.
\begin{definition}[Pair-Wise Probability Gap]
    Given $(h,s)\in [H]\times \mathcal{S}$, the pair-wise probability gap $\overline{\Delta}^M_h(s,a,a')$ for a human comparison of two trajectory sets for action $a$ and $a'$ with cardinality both being $M$ is defined as:
    \begin{align*}
        \overline{\Delta}^M_h(s,a, a') = \underbrace{\mathbb{P}\left( \left. \sum_{m=1}^M f(\tau^m_0) > \sum_{m=1}^M f(\tau^m_1) \right| \tau_0^m \sim \{a_h = a, \pi^*\},~ \tau_1^m \sim \{a_h = a', \pi^*\} \right)}_{\overline{p}_h^M(s,a, a')} - \frac{1}{2},
    \end{align*}
    where the traces $\{\tau^1_0, \tau^2_0, \cdots, \tau^M_0\}$ are independently sampled starting from state $(h,s)$ using immediate action $a_h = a$ and the optimal policy $\{\pi^*_k\}_{k=h+1}^H$ afterwards, while $\{\tau^1_1, \tau^2_1, \cdots, \tau^M_1\}$ are independently sampled starting from state $(h,s)$ using immediate action $a_h = a'$ and the optimal policy $\{\pi^*_k\}_{k=h+1}^H$ afterwards.
\end{definition} 
Here, $\overline{p}_h^M(s,a, a')$ is called the pair-wise preference probability. Note that the pair-wise probability gap can be negative. Notice that the $\mathsf{B\text{-}RUCB}$ routine in Alg.~\ref{alg:BRUCB} relies on the construction of upper and lower confidence intervals using the bonus term $b_h(s,a,a')$ for each action pair, we define the following concentration event which states that the pair-wise preference probability is inside the confidence interval for all states and episodes:
\begin{align*}
    \mathcal{E}_{\mathrm{conc}} = \left\{ \forall (h,s,a, a', k), ~ \overline{p}_h^M(s,a,a') \in [\underbrace{\hat{\sigma}_h(s,a,a') - b_h(s,a,a')}_{l_h(s,a,a')}, \underbrace{\hat{\sigma}_h(s,a,a') + b_h(s,a,a')}_{u_h(s,a,a')}]\right\}.
\end{align*}
We first start with the existence of Condorcet winner, and then prove the sample complexity bound, where we first prove that $\mathcal{E}_{\mathrm{conc}}$ holds with probability $1-\mathcal{O}(\delta)$ (Lemma.~\ref{lemma:conc}). Then, we show that on event $\mathcal{E}_{\mathrm{conc}}$, the crucial lemmas (Lemma.~\ref{lemma:sc_episodic_state} and Lemma.~\ref{lemma:rewardfree_episodic}) holds. Then Theorem.~\ref{thm:sc_episodic} easily follows.
\begin{lemma}\label{lemma:conc}
    Fix $M$ and any $\delta >0$, we have with probability at least $1-\mathcal{O}(\delta)$ that $\mathcal{E}_{\mathrm{conc}}$ holds.
    Moreover, on this event, $\forall (h,s,k), \hat{\pi}_h(s) = \pi^*_h(s)$ as long as it is not empty.
\end{lemma}

\subsection{Proof of Lemma.~\ref{lemma:condorcet}}
Fix a state $(h,s)$ and fix $M$, let $a^* = \pi_h^*(s)$. If $a^*$ is the Condorcet winner, then for any action $a$, we have $\overline{\Delta}_h^M(s,a) > 0$, which requires $\overline{p}_h^M(s,a)> 0.5$. Then, we use concentration to upper bound the probability $1-\overline{p}_h^M(s,a)$ as follows:
\begin{align*}
    1-\overline{p}_h^M(s,a) =& \mathbb{P}\left( \left. \sum_{m=1}^M f(\tau^m_0) < \sum_{m=1}^M f(\tau^m_1) \right| \tau_0^m \sim \pi^*,~ \tau_1^m \sim \{a_h = a, \pi^*\} \right)\\
    =& \mathbb{P}\left( \left. \frac{1}{M}\sum_{m=1}^M [f(\tau^m_0) - f(\tau^m_1)] - \Delta_h(s,a)  < - \Delta_h(s,a) \right| \tau_0^m \sim \pi^*,~ \tau_1^m \sim \{a_h = a, \pi^*\} \right)\\
    \leq & \exp\left( - M\frac{\Delta_h(s,a)^2}{2D^2} \right).
\end{align*}
Then, as long as $M \geq 8 D^2 / \Delta_h(s,a)^2$, we have $1-\overline{p}_h^M(s,a) \leq \exp(-4) < 0.5$, which proves Lemma.~\ref{lemma:condorcet}.

\subsection{Proof of Theorem.~\ref{thm:sc_episodic}}
The proof of Theorem.~\ref{thm:sc_episodic} relies on Lemma.~\ref{lemma:sc_episodic_state}, Lemma.~\ref{lemma:rewardfree_episodic}, and Lemma.~\ref{lemma:conc}. 

First, the correctness of $\hat{\pi}$ follows from Lemma.~\ref{lemma:conc}. We notice two facts on event $\mathcal{E}_{\mathrm{conc}}$: (i) for any state $(h,s,a)$ and any episode $k$, the already identified policy $\hat{\pi}_h(s)$ is either empty or the optimal policy $\pi^*_h(s)$ according to Lemma.~\ref{lemma:conc}; (ii) at the time where the $\mathsf{BSAD}$ algorithm terminates, $\hat{\pi}_h(s)$ for all states $(h,s)$ are not empty, or otherwise the algorithm will not stop according to the design of stopping rule. Therefore, we conclude that at the time when $\mathsf{BSAD}$ terminates, we have $\hat{\pi}_h(s) = \pi^*_h(s)$ for all $(h,s)$. Since $\mathcal{E}_{\mathrm{conc}}$ holds with probability $1-\mathcal{O}(\delta)$, we conclude the $\delta$-PAC property of $\mathsf{BSAD}$.

Second, we show the sample complexity based on Lemma.~\ref{lemma:sc_episodic_state}, Lemma.~\ref{lemma:rewardfree_episodic}. Suppose we have in the current episode $l=h$ and $k = K_h$, by Lemma.~\ref{lemma:rewardfree_episodic}, with probability $1-\mathcal{O}(\delta)$,
\begin{align*}
    K_h \leq \max_s\mathcal{O}\left(\frac{SAh^2 M_h(s)}{\max_\pi p^\pi_h(s)^2}\right) \leq \mathcal{O}\left( \frac{SA^3h^2M \log(\frac{1}{\delta})}{\min_{s,a} \max_\pi [\overline{\Delta}_h^M(s,a) p^\pi_h(s)]^2} \right),
\end{align*}
where the second inequality follows from Lemma.~\ref{lemma:sc_episodic_state}. This shows that the number of episodes with $l=h$ is upper bounded and therefore the total number of episodes is upper bounded by the sum of the RHS as follows:
\begin{align*}
    K \leq \mathcal{O}\left( \sum_{h=1}^H\frac{SA^3h^2M \log(\frac{1}{\delta})}{\min_{s,a} \max_\pi [\overline{\Delta}_h^M(s,a) p^\pi_h(s)]^2} \right).
\end{align*}

\subsection{Proof of Lemma.~\ref{lemma:conc}}
We introduce several notations: we define $\mathcal{E}_{h,s}^{a,a'}(k)$ to be the out-of-concentration event at time $k$ for stage $h$, state $s$, and any two actions $a$ and $a'$:
\begin{align*}
    \mathcal{E}_{h,s}^{a,a'}(k) = \left\{ \overline{p}_h^M(s,a,a') \notin [l_h^k(s,a,a'), u_h^k(s,a,a')] \right\}.
\end{align*}
Notice that $\mathcal{E}_{h,s}^{a,a'}(k) = \mathcal{E}_{h,s}^{a',a}(k)$ since $\overline{p}_h^M(s,a,a') = 1- \overline{p}_h^M(s,a',a)$, $l_h^k(s,a,a') = 1-u_h^k(s,a',a)$, and $u_h^k(s,a,a') = 1-l_h^k(s,a',a)$. Then, we define the following out-of-concentration event:
\begin{align*}
    \mathcal{E}_{h,s}^{a,a'} = \bigcup_{k}\mathcal{E}_{h,s}^{a,a'}(k),\quad
    \mathcal{E}_{h,s} = \bigcup_{k,a,a'}\mathcal{E}_{h,s}^{a,a'}(k), \quad
    \mathcal{E}_{h} = \bigcup_{k,s,a,a'}\mathcal{E}_{h,s}^{a,a'}(k),\quad
    \mathcal{E} = \bigcup_{k,h,s,a,a'}\mathcal{E}_{h,s}^{a,a'}(k).
\end{align*}
Notice that $\mathcal{E}_{\mathrm{conc}} = \mathcal{E}^{\complement}$, the lemma is equivalent to show $\prob(\mathcal{E}) = \mathcal{O}(\delta)$. We first use the set relationship to decompose the event as follows:
\begin{align*}
    \prob(\mathcal{E}) =& \prob\left( \bigcup_{h=1}^H \mathcal{E}_{h} \right)
    = \prob\left( \mathcal{E}_{H} \right) + \prob\left( \left(\bigcup_{h=1}^{H-1} \mathcal{E}_{h} \right) \cap \mathcal{E}_{H}^{\complement} \right)\\
    =& \prob\left( \mathcal{E}_{H} \right) + \prob\left( \mathcal{E}_{H-1} \cap \mathcal{E}_{H}^{\complement} \right) + \prob\left( \left(\bigcup_{h=1}^{H-2} \mathcal{E}_{h} \right) \cap \left(\bigcap_{h=H-1}^H \mathcal{E}_h^{\complement} \right) \right)\\
    =& \sum_{h=1}^H \prob\left( \mathcal{E}_h \cap \left(\bigcap_{i=h+1}^H \mathcal{E}_i^{\complement} \right) \right).
\end{align*}
We then proceed to bound each term in the summation one by one following a backward induction argument. Note that the $\mathsf{B\text{-}RUCB}$ sub-routine will not be called for stage $h$ unless all stages after $h$ has identified $\hat{\pi}_h(s)$. We start with stage $H$ and bound $\prob(\mathcal{E}_H)$. We first use a union bound:
\begin{align*}
    \prob\left( \mathcal{E}_H\right) 
    =& \prob\left( \bigcup_{k,s,a,a'}\mathcal{E}_{H,s}^{a,a'}(k) \right)
    \leq \sum_{k,s,a,a'} \prob\left( \mathcal{E}_{H,s}^{a,a'}(k) \right)\\
    =& \sum_{k,s,a,a'} \prob\left(  \overline{p}_H^M(s,a,a') \notin [l_H^k(s,a,a'), u_H^k(s,a,a')]  \right)\\
    \leq & \sum_{k,s,a,a'} \left[ \prob\left( \overline{p}_H^M(s,a,a') < l_H^k(s,a,a')  \right) + \prob\left( \overline{p}_H^M(s,a,a') > u_H^k(s,a,a') \right) \right].
\end{align*}
Recall that at time stage $H$, the reward $f(\{(s, a)\}_{H:H})$ is deterministic for all state $s$ and action $a$. Therefore, the wining probability $\overline{p}_H^M(s,a,a')$ is merely an indicator function:
\begin{align*}
    \overline{p}_H^M(s,a,a') = \mathbbm{1}\left\{f(\{(s, a)\}_{H:H}) > f(\{(s, a')\}_{H:H}) \right\}.
\end{align*}
Moreover, the estimation statistics of the wining probability $\hat{\sigma}_H^k(s,a,a')$ also takes only two values $1$ or $0$, since the reward is deterministic, i.e.,
\begin{align*}
    \hat{\sigma}_H^k(s,a,a') =& \frac{w_H^k(s,a,a')}{N_H^k(s,a,a')} = \mathbbm{1}\left\{f(\{(s, a)\}_{H:H}) > f(\{(s, a')\}_{H:H}) \right\}.
\end{align*}
Thus we conclude that $\overline{p}_H^M(s,a,a') = \hat{\sigma}_H^k(s,a,a')$. Now, we can simplify and bound the two probabilities as follows:
\begin{align*}
    \prob\left( \overline{p}_H^M(s,a,a') < \hat{\sigma}_H^k(s,a,a') - b_H^k(s,a,a')  \right) 
    = \prob\left( b_H^k(s,a,a') < 0  \right)  =0,\\
    \prob\left( \overline{p}_H^M(s,a,a') > \hat{\sigma}_H^k(s,a,a') + b_H^k(s,a,a') \right)
    = \prob\left( b_H^k(s,a,a') < 0  \right)  =0.
\end{align*}
And thus we have $\prob\left( \mathcal{E}_H\right) = 0$. Next, we bound $\prob( \mathcal{E}_h \cap (\bigcap_{i=h+1}^H \mathcal{E}_i^{\complement} ))$ for a fixed stage $h$, which supposes the concentration event $\mathcal{E}_i^{\complement}$ holds for all $i\geq h+1$. For simplicity, we use $\tilde{\mathcal{E}}_h^{\complement}$ to denote $\bigcap_{i=h+1}^H \mathcal{E}_i^{\complement}$. We also first use union bound to decompose the probability as follows:
\begin{align*}
    \prob\left( \mathcal{E}_h \cap \tilde{\mathcal{E}}_h^{\complement}\right)
    =& \sum_{a,a',s}\prob\left(\{\exists k, \mathcal{E}_h^{a,a'}(k)\} \cap \tilde{\mathcal{E}}_h^{\complement}\right)\\
    =& \sum_{a,a',s}\prob\left(\left\{\exists k,  \overline{p}_h^M(s,a,a') \notin [l_h^k(s,a,a'), u_h^k(s,a,a')] \right\} \cap \tilde{\mathcal{E}}_h^{\complement}\right)\\
    =& \sum_{a,a',s}\prob\left(\left\{\exists k,  \overline{p}_h^M(s,a,a') <l_h^k(s,a,a')\right\} \cap \tilde{\mathcal{E}}_h^{\complement}\right) + \prob\left(\left\{\exists k,  \overline{p}_H^M(s,a,a') >u_h^k(s,a,a')\right\} \cap \tilde{\mathcal{E}}_h^{\complement}\right).
\end{align*}
We first analyze the first probability term. Let $\tilde{w}_h^n(s,a,a')$ be the number of wins for action $a$ against $a'$ in their first $n$ comparisons, we have:
\begin{align*}
    &\prob\left(\left\{\exists k,  \overline{p}_h^M(s,a,a') <l_h^k(s,a,a')\right\} \cap \tilde{\mathcal{E}}_h^{\complement}\right) \\
    = & \prob\left(\left\{\exists k,  \overline{p}_h^M(s,a,a') <\hat{\sigma}_h^k(s,a,a') - b_h^k(s,a,a')\right\} \cap \tilde{\mathcal{E}}_h^{\complement}\right)\\
    =& \prob\left(\left\{\exists k,  \overline{p}_h^M(s,a,a') <\frac{w_h^k(s,a,a')}{N_h^k(s,a,a')} - \sqrt{\frac{c\log\left( SAHk/\delta \right)}{N_h^k(s,a,a')}}\right\} \cap \tilde{\mathcal{E}}_h^{\complement}\right)\\
    \leq & \prob\left(\left\{\exists k,  \overline{p}_h^M(s,a,a') <\frac{w_h^k(s,a,a')}{N_h^k(s,a,a')} - \sqrt{\frac{c\log\left( SAHN_h^k(s,a,a')/\delta \right)}{N_h^k(s,a,a')}}\right\} \cap \tilde{\mathcal{E}}_h^{\complement}\right)\\
    \leq & \prob\left(\left\{\exists n,  \overline{p}_h^M(s,a,a') <\frac{\tilde{w}_h^n(s,a,a')}{n} - \sqrt{\frac{c\log\left( SAHn/\delta \right)}{n}}\right\} \cap \tilde{\mathcal{E}}_h^{\complement}\right)\\
    \leq & \sum_{n} \prob\left(\left\{\overline{p}_h^M(s,a,a') <\frac{\tilde{w}_h^n(s,a,a')}{n} - \sqrt{\frac{c\log\left( SAHn/\delta \right)}{n}}\right\} \cap \tilde{\mathcal{E}}_h^{\complement}\right),
\end{align*}
where the first inequality is due to $k\geq N_h^k(s,a,a')$, and the second inequality is by transferring from counting the number of episodes $k$ to the number of pulls $n$. The last inequality is due to a union bound. Conditioned on this event, notice that $\tilde{w}_h^n(s,a,a') = \sum_{i=1}^n \sigma_i$ and each $\sigma_i$ is the $i$-th comparison result between the following two policies for $M$ trajectories:
\begin{align*}
    &\left\{ a_h = a, a_{h+1} = \hat{\pi}_{h+1}(s_{h+1}), \cdots, a_H = \hat{\pi}_H(s_H) \right\},\\
    &\left\{ a_h = a', a_{h+1} = \hat{\pi}_{h+1}(s_{h+1}), \cdots, a_H = \hat{\pi}_H(s_H) \right\}.
\end{align*}
It is somewhat difficult to analyze the statistics of $\sigma_i$ since $\hat{\pi}$ is trajectory dependent. However, we will be utilizing the next lemma which shows that the algorithm is mistake-free on any stage $h$ if the concentration event $\mathcal{E}_h^\complement$ holds. 
\begin{lemma}\label{lemma:conc-optarm}
    For each stage $h$ and concentration event $\mathcal{E}_h^{\complement}$, we have:
    \begin{align*}
        \mathcal{E}_h^{\complement} \subset \left\{ \forall(k,s), \hat{\pi}_h(s) = \pi^*_h(s) \right\}.
    \end{align*}
\end{lemma}
Therefore, on event $\tilde{\mathcal{E}}_h^{\complement}$, the output policy $\hat{\pi}_{h+1}, \cdots, \hat{\pi}_H$ is equivalent to $\pi^*_{h+1}, \cdots, \pi^*_H$. And therefore, $\tilde{w}^n_h(s,a,a')/n$ is by definition an unbiased estimator of the wining probability $\overline{p}_h^M(s,a,a')$. Then, we can use Azuma-Hoeffding's inequality as follows:
\begin{align*}
    \prob\left(\left\{\overline{p}_h^M(s,a,a') <\frac{\tilde{w}_h^n(s,a,a')}{n} - \sqrt{\frac{c\log\left( SAHn/\delta \right)}{n}}\right\} \cap \tilde{\mathcal{E}}_h^{\complement}\right) \leq \left(\frac{\delta}{SAHn}\right)^{\frac{c}{2}}.
\end{align*}
We can thus bound the probability large deviation event assuming $c$ is large enough:
\begin{align*}
    \prob\left(\left\{\exists k,  \overline{p}_h^M(s,a,a') <l_h^k(s,a,a')\right\} \cap \tilde{\mathcal{E}}_h^{\complement}\right) \leq \sum_n \left(\frac{\delta}{SAHn}\right)^{\frac{c}{2}} \leq \frac{\delta}{SA^2H} .
\end{align*}
Similarly, the other term can also be bounded. So, we have bound on the out-of-concentration event:
\begin{align*}
    \prob(\mathcal{E}) = \sum_{h=1}^H \prob\left( \mathcal{E}_h \cap \left(\bigcap_{i=h+1}^H \mathcal{E}_i^{\complement} \right) \right)
    \leq \sum_{h=1}^H \sum_{a,a',s} \mathcal{O}\left( \frac{\delta}{SA^2H}\right) = \mathcal{O}(\delta).
\end{align*}
And also by Lemma.~\ref{lemma:conc-optarm}, we know that:
\begin{align*}
    \mathcal{E}_{\mathrm{conc}} = \mathcal{E}^{\complement} \subset \left\{ \forall (h,s,k), \hat{\pi}_h(s) = \pi^*_h(s) \right\}.
\end{align*}

\subsubsection{Proof of Lemma.~\ref{lemma:conc-optarm}}
We first recall that $\mathcal{E}_h^{\complement}$ is the ``good'' concentration event:
\begin{align*}
    \mathcal{E}_h^{\complement} = \left\{ \forall (k,s,a,a'), ~ \overline{p}_h^M(s,a,a') \in [l_h^k(s,a,a'), u_h^k(s,a,a')] \right\}.
\end{align*}
Suppose there exists a state $s$ such that $\hat{\pi}_h(s) = \tilde{a} \neq \pi^*_h(s) = a^*$ after the stage $h$ has been iterated, i.e., $l<h$. There must have existed a time $k$ such that:
\begin{align*}
    l_h^k(s,\tilde{a},a^*) = \hat{\sigma}_h^k(s,\tilde{a},a^*) - b_h^k(s,\tilde{a},a^*) >\frac{1}{2}.
\end{align*}
Since on event $\mathcal{E}_h^{\complement}$, we have $l_h^k(s,\tilde{a},a^*) \leq \overline{p}_h^M(s,\tilde{a},a^*)$ for all $(h,s)$ and for any actions $(\tilde{a}, a^*)$, it implies:
\begin{align*}
    \overline{p}_h^M(s,\tilde{a},a^*) >\frac{1}{2},
\end{align*}
which is contradictory to the fact that $a^*$ is a Condorcet winner in this state. Therefore, we conclude that:
\begin{align*}
    \mathcal{E}_h^{\complement} \subset \left\{ \forall(k,s), \hat{\pi}_h(s) = \pi^*_h(s) \right\}.
\end{align*}

\subsection{Proof of Lemma.~\ref{lemma:sc_episodic_state}}
We will show that Lemma.~\ref{lemma:sc_episodic_state} holds on the concentration event $\mathcal{E}_{\mathrm{conc}}$. Therefore, in this section, we assume $\mathcal{E}_{\mathrm{conc}}$ is true. We first rely on the following lemma to bound the number of comparisons (human expert queries) between two actions for a given $(h,s)$:
\begin{lemma}\label{lemma:comparison_2arm}
    Fix a stage $h$ and state $s$ and on $\mathcal{E}_{\rm{conc}}$, the number of comparisons between two sub-optimal actions $a$ and $a'$ is bounded for all episode $k$ before the criteria $\mathcal{T}_h(s)$ is met:
    \begin{align*}
        N_h^k(s,a,a') \leq 
            \mathcal{O}\left( \frac{ \log\left(\frac{SAHk}{\delta} \right) }{ \min\left\{\overline{\Delta}_h^M(s,a)^2, \overline{\Delta}_h^M(s,a')^2 \right\}}  \right).
    \end{align*}
    The number of comparisons between a sub-optimal action $a$ and the optimal action $\pi^*_h(s)$ is bounded for all episode $k$:
    \begin{align*}
        N_h^k\left(s,a,\pi^*_h(s)\right) \leq 
            \mathcal{O}\left( \frac{ \log\left(\frac{SAHk}{\delta} \right)  }{\overline{\Delta}_h^M(s,a)^2} \right).
    \end{align*}
\end{lemma}
With this lemma in hand, we only focus on the traces that visits stage $(h,s)$, and each $2M$ visit will correspond to $1$ comparisons as defined by the algorithm, so we have the number of visits bounded by the total number of comparisons multiplied by $2M$ as follows:
\begin{align*}
    M_h(s) \leq& 2M \sum_{a,a'} N_h^k(s,a,a') = \tilde{\mathcal{O}}\left( \sum_{a\neq \pi^*_h(s)} \frac{ M\log\left(\frac{1}{\delta} \right)  }{\overline{\Delta}_h^M(s,a)^2}  +   \sum_{ a\neq a' }^{a,a'\neq \pi_h^*(s)}  \frac{ M \log\left(\frac{1}{\delta} \right) }{ \min\left\{\overline{\Delta}_h^M(s,a)^2, \overline{\Delta}_h^M(s,a')^2 \right\}}     \right).
\end{align*}
We re-order the actions on this $(h,s)$ according to the probability gap to obtain a permutation $\{a_1, a_2, \cdots, a_A\}$ such that $a_1$ is the optimal action and $\overline{\Delta}_h^M(s,a_2)\leq \overline{\Delta}_h^M(s,a_2)\leq \cdots, \overline{\Delta}_h^M(s,a_A)$. Notice that a term corresponding to $a_1$ will not appear, a term involving $a_2$ will only appear once in the first summation, while a term corresponding to $a_3$ will appear twice, once in the first summation and once in the second when compared to $a_2$, and so on. Therefore, we can rewrite the RHS as follows:
\begin{align*}
    \tilde{\mathcal{O}}\left( \sum_{i=2}^A \frac{i}{\overline{\Delta}_h^M(s,a_i)^2} M\log\left(\frac{1}{\delta} \right) \right)
        = \tilde{\mathcal{O}}\left( \frac{A^2 M \log\left(\frac{1}{\delta} \right)  }{\min_{a}\overline{\Delta}_h^M(s,a)^2} \right).
\end{align*}
Remark: here we neglect the dependence on $k$ since $k$ is smaller than the total sample complexity bound for identifying the optimal action in step $h$, which is at most polynomial to the $S$, $A$, $H$, and $\log(\delta^{-1})$. So overall $\log(k)$ contributes to a logarithmic term and is hidden in the $\tilde{\mathcal{O}}$ notation.

\subsubsection{Proof of Lemma.~\ref{lemma:comparison_2arm}}
We assume the concentration event $\mathcal{E}_{\mathrm{conc}}$ holds, which happens with probability at least $1-\mathcal{O}(\delta)$. On this event, we can conclude that for all episode $k$, and for all $(h,s)$, set $\mathcal{C}_h(s)$ in the $\mathsf{B\text{-}RUCB}$ sub-routine (Alg.~\ref{alg:BRUCB}, Line 3) is not empty, since we have:
\begin{align*}
    \hat{\sigma}_h\left(s,\pi^*_h(s), a'\right) + b_h^k\left(s,\pi^*_h(s), a'\right) \geq \overline{p}_h^M\left(s,\pi^*_h(s), a'\right) \geq \frac{1}{2},
\end{align*}
for all other actions $a'$. Suppose we now bound the number of comparisons between a sub-optimal action $a$ and the optimal action $\pi^*_h(s)$ for a fixed state and stage $(h,s)$. For any episode $k$, in which these two actions are chosen for comparison, we analyze the following two cases:

\textbf{Case 1:} Suppose $\hat{a} = \pi^*_h(s)$ and $\tilde{a} = a$. We must have:
\begin{align*}
    \exists a'\neq \pi^*_h(s), ~ \hat{\sigma}_h^k\left(s,\pi^*_h(s), a'\right) -  b_h^k\left(s,\pi^*_h(s),a'\right) \leq \frac{1}{2},
\end{align*}
or otherwise, the optimal action is already found for this state before this episode starts. This requirement is equivalent to:
\begin{align*}
    \exists a' \neq \pi^*_h(s), ~ \hat{\sigma}_h^k\left(s,a',\pi^*_h(s)\right) +  b_h^k\left(s,a',\pi^*_h(s)\right) \geq \frac{1}{2},
\end{align*}
If action $a$ does not satisfy this requirement, it is not difficult to see that the upper confidence bound of action $a'$ is larger than action $a$, which action $a$ cannot be picked by the algorithm, i.e.,
\begin{align*}
    \hat{\sigma}_h^k\left(s,a',\pi^*_h(s)\right) +  b_h^k\left(s,a',\pi^*_h(s)\right)\geq \frac{1}{2} \geq \hat{\sigma}_h^k\left(s,a,\pi^*_h(s)\right) +  b_h^k\left(s,a,\pi^*_h(s)\right).
\end{align*}
If action $a$ happens to satisfy this requirement, we have:
\begin{align*}
    \hat{\sigma}_h^k\left(s,a, \pi^*_h(s) \right) + b_h^k\left(s,a,\pi^*_h(s) \right) \geq \frac{1}{2}.
\end{align*}
Notice that on the concentration event $\mathcal{E}_{\mathrm{conc}}$, we have:
\begin{align*}
    \overline{p}_h^M\left(s,a,\pi^*_h(s) \right) +  2b_h^k\left(s,a,\pi^*_h(s)\right) \geq \hat{\sigma}_h^k\left(s,a, \pi^*_h(s) \right) +  b_h^k\left(s,a,\pi^*_h(s)\right) \geq \frac{1}{2}.
\end{align*}
This induces:
\begin{align*}
    \sqrt{\frac{ c \log\left(\frac{ASHk}{\delta}\right)}{N_h^k(s,a, \pi^*_h(s))}}\geq \frac{\overline{\Delta}_h^M(s,a)}{2},
\end{align*}
which induces:
\begin{align*}
    N_h^k\left(s,a,\pi^*_h(s)\right) \leq 
        \mathcal{O}\left( \frac{ \log\left(\frac{SAHk}{\delta} \right)  }{\overline{\Delta}_h^M(s,a)^2} \right).
\end{align*}

\textbf{Case 2:} Suppose $\hat{a} = a$ and $\tilde{a} = \pi^*_h(s)$. We must have $a\in\mathcal{C}$, which leads to:
\begin{align*}
    \hat{\sigma}_h^k\left(s,a, \pi^*_h(s) \right) + b_h^k\left(s,a,\pi^*_h(s) \right) \geq \frac{1}{2}.
\end{align*}
Following the same argument, we have:
\begin{align*}
    N_h^k\left(s,a,\pi^*_h(s)\right) \leq 
        \mathcal{O}\left( \frac{ \log\left(\frac{SAHk}{\delta} \right)  }{\overline{\Delta}_h^M(s,a)^2} \right).
\end{align*}

Now, suppose neither action $a$ nor $a'$ is the optimal action, for any episode $k$, in which the two actions are chosen for comparison, we can still separate in two cases:

\textbf{Case 1:} Suppose $\hat{a} = a$ and $\tilde{a} = a'$. We must have $a\in\mathcal{C}$ which infers:
\begin{align*}
    \hat{\sigma}_h^k(s,a,a') + b_h^k(s,a,a') \geq \frac{1}{2}, \quad \text{equivalently} \quad \hat{\sigma}_h^k(s,a',a) - b_h^k(s,a',a) \leq \frac{1}{2}.
\end{align*}
On the other hand, since $a'$ has the largest upper confidence bound, we have on the concentration event that:
\begin{align*}
    \hat{\sigma}_h^k(s,a',a) + b_h^k(s,a',a) \geq \hat{\sigma}_h^k(s,\pi^*(s),a) + b_h^k(s,\pi^*(s),a)\geq \overline{p}_h^M(s,\pi^*(s),a)
\end{align*}
Combining the two inequalities, we have:
\begin{align*}
    2b_h^k(s,a',a) = 2\sqrt{\frac{ c \log\left(\frac{ASHk}{\delta}\right)}{N_h^k(s,a, a')}} \geq \overline{\Delta}_h^M(s,a).
\end{align*}
This induces:
\begin{align*}
    N_h^k(s,a, a')\leq \mathcal{O}\left( \frac{ \log\left(\frac{SAHk}{\delta} \right)  }{\overline{\Delta}_h^M(s,a)^2} \right).
\end{align*}

\textbf{Case 2:} Suppose the other way around that $\hat{a} = a'$ and $\tilde{a} = a$. With the same argument, we have:
\begin{align*}
    N_h^k(s,a, a')\leq \mathcal{O}\left( \frac{ \log\left(\frac{SAHk}{\delta} \right)  }{\overline{\Delta}_h^M(s,a')^2} \right).
\end{align*}
Therefore, we conclude that:
\begin{align*}
    N_h^k(s,a,a') \leq 
        \mathcal{O}\left( \frac{ \log\left(\frac{SAHk}{\delta} \right) }{ \min\left\{\overline{\Delta}_h^M(s,a)^2, \overline{\Delta}_h^M(s,a')^2 \right\}}  \right).
\end{align*}

\subsection{Proof of Lemma.~\ref{lemma:rewardfree_episodic}}
Similarly, in this proof, we assume $\mathcal{E}_{\mathrm{conc}}$ holds. The proof resembles the proof of \citep[Theorem 3]{zhang2020rewardfree}, but we use a different exploration bonus and need to take care of the termination step. For notation, we adopt the same quantities defined in \citep[(4.2)]{jin2018qlearning}:
\begin{align*}
    \alpha_t^0 = \Pi_{j=1}^t(a-\alpha_j), \quad \alpha_t^i = \alpha_i \Pi_{j=i+1}^t(a-\alpha_j).
\end{align*}
The following lemma summarizes the properties of $\alpha_t^i$ which is proved in~\citep{jin2018qlearning}:
\begin{lemma}\label{lemma:alpha}
    The following properties hold for $\alpha_t^i$:
    \begin{enumerate}
        \item $\sum_{i=1}^t\alpha_t^i = 1$ and $\alpha_t^0 = 0$ for $t\geq 1$; $\sum_{i=1}^t\alpha_t^i = 0$ and $\alpha_t^0 = 1$ for $t=0$.
        \item $\frac{1}{\sqrt{t}}\leq \sum_{i=1}^t \frac{\alpha_t^i}{\sqrt{i}}\leq \frac{2}{\sqrt{t}}$ for $t\geq 1$.
        \item $\sum_{i=1}^t(\alpha_t^i)^2 \leq \frac{2H}{t}$ for $t\geq 1$.
        \item $\sum_{t=i}^{\infty}\alpha_t^i = 1+\frac{1}{H}$ for every $i\geq 1$.
    \end{enumerate}
\end{lemma}

Fix $l$, $\forall h \leq l-1$, suppose in the current episode, we have $t = L_h^k(s,a)$. By the update rule of the $J_h(s,a)$ function in our reward-free exploration, and according to \citep[(4.3)]{jin2018qlearning}, suppose $(s,a)$ is taken at step $h$ in episodes with $k=k_1, \cdots, k_t<k$. We use superscripts $k$ to denote the episode index, and we have:
\begin{align*}
    J_h^k(s,a) = \alpha_t^0 + \sum_{i=1}^t \alpha_t^i[W_{h+1}^{k_i}(s_{h+1}^{k_i}) + b_i]
    \geq \sum_{i=1}^t \alpha_t^i b_i
    \geq \sum_{i=1}^t \alpha_t^i b_t = b_t,
\end{align*}
where the first inequality is by the positivity of the value function and the second inequality is by the monotonicity of $b_i$. The last equality is due to the first property of Lemma.~\ref{lemma:alpha}. We have the following two situations:

\textbf{Case 1:} $\forall h\leq l-2$, we have:
\begin{align*}
    J_h^k(s,a) =& \alpha_t^0 + \sum_{i=1}^t \alpha_t^i[W_{h+1}^{k_i}(s_{h+1}^{k_i}) + b_i] \\
    =& \alpha_t^0 + \sum_{i=1}^t \alpha_t^i[W_{h+1}^{k_i}(s_{h+1}^{k_i}) - \expt[W_{h+1}^{k_i}(s')|s,a] + \expt[W_{h+1}^{k_i}(s')|s,a] + b_i] \\
    =& \alpha_t^0 + \sum_{i=1}^t \alpha_t^i \expt[W_{h+1}^{k_i}(s')|s,a] + \sum_{i=1}^t \alpha_t^i[W_{h+1}^{k_i}(s_{h+1}^{k_i}) - \expt[W_{h+1}^{k_i}(s')|s,a]] +  \sum_{i=1}^t \alpha_t^i b_i \\
    \geq & \sum_{i=1}^t \alpha_t^i \expt[W_{h+1}^{k_i}(s')|s,a] - \sqrt{\frac{H\iota}{t}} + b_t\\
    = & \sum_{i=1}^t \alpha_t^i \sum_{s'}P_h\left(s'|s,a \right) W_{h+1}^{k_i}(s'),
\end{align*}
where the first inequality uses Azuma-Hoeffding's inequality as in \citep[(4.4)]{jin2018qlearning}, and holds with probability $1-\mathcal{O}(\delta / S^2AH)$.

\textbf{Case 2:} $\forall h = l-1$, we have:
\begin{align*}
    J_{l-1}^k(s,a) = \alpha_t^0 + \sum_{i=1}^t \alpha_t^i[W_l^{k_i}(s_l^{k_i}) + b_i] \geq \sum_{i=1}^t \alpha_t^i \sum_{s'}P_{l-1}\left(s'|s,a \right) \min\{b_l^{k_i}(s'), 1\},
\end{align*}
where $b_l^{k_i}$ is the value function in step $l$ at the episode with $k=k_i$. Then, for any fixed $s^*$ in stage $l$, we iterate as follows:
\begin{align*}
    J_{l-1}^k(s,a) \geq& \sum_{i=1}^t \alpha_t^i \sum_{s'}P_{l-1}\left(s'|s,a \right) \min\{b_l^{k_i}(s'), 1\}
    \geq \sum_{i=1}^t \alpha_t^i P_{l-1}\left(s^*|s,a \right) \min\{b_l^{k_i}(s^*), 1\}\\
    = & P_{l-1}\left(s^*|s,a \right) \min\{b_l^{k}(s^*), 1\},
\end{align*}
where the last inequality uses Lemma.~\ref{lemma:alpha} again. For $h< h'$, define $p_{h,h'}^\pi(s'|s)$ to be the transition probability starting from $(h,s)$ and ending in $(h',s')$ following policy $\pi$, i.e.,
\begin{align*}
    p_{h,h'}^\pi(s'|s) = \prob(s_h' = s' | s_h = s, \pi).
\end{align*}
Therefore, we have:
\begin{align*}
    W_{l-1}^k(s) =& \max_a J_{l-1}^k(s,a) \geq \max_a P_{l-1}\left(s^*|s,a \right) \min\{b_l^{k}(s^*), 1\}\\
    =& \max_\pi p^{\pi}_{l-1, l}(s^*|s)\min\{b_l^{k}(s^*), 1\}
\end{align*}
Now we follow induction to show that for any $h$, $W_h^k(s) \geq \max_\pi p^{\pi}_{h, l}(s^*|s)\min\{b_l^{k}(s^*), 1\}$. Suppose this is satisfied with $h+1$. We have:
\begin{align*}
    J_h^k(s,a) \geq& \sum_{i=1}^t \alpha_t^i \sum_{s'}P_h\left(s'|s,a \right) W_{h+1}^{k_i}(s')\\
    \geq & \sum_{i=1}^t \alpha_t^i \sum_{s'}P_h\left(s'|s,a \right) \max_\pi p^{\pi}_{h+1, l}(s^*|s)\min\{b_l^{k_i}(s^*), 1\} \\
    \geq & \sum_{s'}P_h\left(s'|s,a \right) \max_\pi p^{\pi}_{h+1, l}(s^*|s)\min\{b_l^{k}(s^*), 1\}.
\end{align*}
Then, we can lower bound the value function as:
\begin{align*}
    W_h^k(s) \geq & \max_a \sum_{s'}P_h\left(s'|s,a \right) \max_\pi p^{\pi}_{h+1, l}(s^*|s)\min\{b_l^{k}(s^*), 1\} \\
    = & \max_\pi p^{\pi}_{h, l}(s^*|s)\min\{b_l^{k}(s^*), 1\}.
\end{align*}
Then, by induction, we established that: 
\begin{align*}
    \expt_{\mu_0}[W_1^k(s_1)] \geq \max_\pi p^{\pi}_l(s^*)\min\{b_l^{k}(s^*), 1\}.
\end{align*}
This also implies for any $K$, we have:
\begin{align*}
    \sum_{k=1}^K \expt_{\mu_0}[W_1^k(s_1)] 
    \geq \sum_{k=1}^K \max_\pi p^{\pi}_l(s^*)\min\{b_l^{k}(s^*), 1\}
    \geq  K\max_\pi p^{\pi}_l(s^*)\min\{b_l^{K}(s^*), 1\}.
\end{align*}
Therefore, we have:
\begin{align*}
    \min\{b_l^{k}(s^*), 1\} \leq \frac{\sum_{k=1}^K \expt_{\mu_0}[W_1^k(s_1)]}{K\max_\pi p^{\pi}_l(s^*)}.
\end{align*}
The rest is to bound $\sum_{k=1}^K \expt_{\mu_0}[W_1^k(s_1)]$, which we use the following lemma:
\begin{lemma}\label{lemma:regret}
Fix $l$, for any $K$ which the stopping rule for step $l$ is not triggered:
    \begin{align*}
        \sum_{k=1}^K \expt_{\mu_0}[W_1^k(s_1)]  = \mathcal{O} \left( SAl +l\sqrt{SAHK\iota} \right).
    \end{align*}
\end{lemma}
Therefore, when $K \geq \max\{SA/\iota, SAl^2\iota/\max_\pi p^{\pi}_l(s^*)^2\}$, we have with probability $1-\mathcal{O}(\delta/S)$:
\begin{align*}
    K\leq  \frac{SAl^2 M_l(s^*)}{\max_\pi p^\pi_l(s^*)^2}.
\end{align*}
Since $s^*$ is arbitrary, an extra use of the union bound will prove the original lemma.

\subsection{Proof of Lemma.~\ref{lemma:regret}}
For $h = l-1$, let $t = L_{l-1}^k(s_{l-1}^k,a_{l-1}^k)$ we have according to \citep[(4.3)]{jin2018qlearning}:
\begin{align*}
    W_{l-1}^k(s_{l-1}^k) \leq & J_{l-1}(s_{l-1}^k, a_{l-1}^k)
    = \alpha_t^0 + \sum_{i=1}^t \alpha_t^i [b_l^{k_i}(s_l^{k_i}) + b_i)]\\
    =& \alpha_t^0 + \sum_{i=1}^t \alpha_t^i b_i + \sum_{i=1}^t \alpha_t^i b_l^{k_i}(s_l^{k_i}).
\end{align*}
We also sum over $k$ for the last term as follows. without ambiguity, we let $L_{l-1}^k = L_{l-1}^k(s_{l-1}^k,a_{l-1}^k)$:
\begin{align*}
    \sum_{k=1}^K\sum_{i=1}^{L_{l-1}^k} \alpha_{L_{l-1}^k}^i b_l^{k_i}(s_l^{k_i}) 
    \leq& \sum_{k'=1}^K b_l^{k'}(s_l^{k'})\sum_{i = L_{l-1}^{k'}+1}^\infty \alpha_i^{t} 
    \leq  \left( 1+\frac{1}{H} \right) \sum_{k=1}^K b_l^k(s_l^k)\\
    = & \left( 1+\frac{1}{H} \right) \sum_{s}\sum_{n_s = 1}^{N_l^K(s)}\sqrt{\frac{H\iota}{n_s}}
    \leq  \left( 1+\frac{1}{H} \right) \sqrt{SHK\iota}
\end{align*}
where in the first inequality we interchange the summation, and in the second inequality, we use Lemma.~\ref{lemma:alpha}. The last inequality uses the Cauchy-Schwarz inequality. Similarly, we can sum over $k$ for the second term as:
\begin{align*}
    \sum_{k=1}^K\sum_{i=1}^{L_{l-1}^k}\alpha_{L_{l-1}^k}^i b_i 
    =& \mathcal{O}(1) \sum_{k=1}^K \sqrt{\frac{H\iota}{L_{l-1}^k(s_{l-1}^k, a_{l-1}^k)}} = \mathcal{O}(1) \sum_{s=1}^S\sum_{a=1}^A\sum_{n_{s,a} = 1}^{N_{l-1}^K(s,a)}\sqrt{\frac{H\iota}{n_{s,a}}}\\
    \leq &\mathcal{O}\left( \sqrt{SAKH\iota} \right),
\end{align*}
where the first equality is due to (4) of Lemma.~\ref{lemma:alpha} and the last inequality uses Cauchy-Schwarz inequality. Notice that by the first property of Lemma.~\ref{lemma:alpha}, we have:
\begin{align*}
    \sum_{k=1}^K \alpha_t^0 = \sum_{k=1}^K \mathbbm{1}\{N_{l-1}^k(s_{l-1}^k, a_{l-1}^k) = 0\} \leq SA.
\end{align*}
Therefore, we can bound the sum over the value function as follows:
\begin{align*}
    \sum_{k=1}^K W_{l-1}^k(s_{l-1}^k) \leq SA + \mathcal{O}\left( \sqrt{SAKH\iota} \right).
\end{align*}

For $h\leq l-2$, similarly, $t = L_{h}^k(s_{h}^k,a_{h}^k)$, and we have according to \citep[(4.3)]{jin2018qlearning}:
\begin{align*}
    W_h^k(s_h^k) \leq &J_h^k(s_h^k, a_h^k) = \alpha_t^0 + \sum_{i=1}^t \alpha_t^i [W_{h+1}^{k_i}(s_{h+1}^{k_i}) + b_i]
    = \alpha_t^0 + \sum_{i=1}^t \alpha_t^i b_i + \sum_{i=1}^t \alpha_t^iW_{h+1}^{k_i}(s_{h+1}^{k_i} ).
\end{align*}
Then, we sum over $k$ and uses Lemma.~\ref{lemma:alpha} as follows:
\begin{align*}
    \sum_{k=1}^K\sum_{i=1}^t \alpha_t^i W_{h+1}^{k_i}(s_{h+1}^{k_i}) \leq  \sum_{k=1}^K W_{h+1}^k(s_{h+1}^k)\sum_{i = t+1}^\infty \alpha_i^{t} = \left( 1+\frac{1}{H} \right) \sum_{k=1}^K W_{h+1}^k(s_{h+1}^k).
\end{align*}
Therefore, we have for $h\leq l-2$:
\begin{align*}
    \sum_{k=1}^K W_h^k(s_h^k) \leq& \sum_{k=1}^K \alpha_t^0 + \sum_{k=1}^K \sum_{i=1}^t \alpha_t^i b_i + \left( 1+\frac{1}{H} \right) \sum_{k=1}^K W_{h+1}^k(s_{h+1}^k) 
\end{align*}
Notice that by the first property of Lemma.~\ref{lemma:alpha}, we have:
\begin{align*}
    \sum_{k=1}^K \alpha_t^0 = \sum_{k=1}^K \mathbbm{1}\{N_h^k(s_h, a_h) = 0\} \leq SA.
\end{align*}
Also by the second property of Lemma.~\ref{lemma:alpha}, we have:
\begin{align*}
    \sum_{k=1}^K \sum_{i=1}^t \alpha_t^i b_i \leq \sum_{k=1}^K \mathcal{O}(1) b_t \leq \mathcal{O}(1)\sum_{s,a}\sum_{i=1}^{N_h^K(s,a)}\sqrt{\frac{H\iota}{i}} = \mathcal{O}\left( \sqrt{SAKH\iota} \right),
\end{align*}
where the last inequality uses Cauchy Schwarz inequality, we have:
\begin{align*}
    \sum_{k=1}^K W_h^k(s_h^k) = & \mathcal{O}\left( SA + \sqrt{SAHK\iota} \right) + \left( 1+\frac{1}{H} \right) \sum_{k=1}^K W_{h+1}^k(s_{h+1}^k) 
\end{align*}
And continue to roll out until step $l$, we have:
\begin{align*}
    \sum_{k=1}^K W_1^k(s_1^k) = \mathcal{O}\left( SAl + l\sqrt{SAHK\iota}\right).
\end{align*}

\section{Detailed Discussion on Generalization to Discounted MDPs}
In this section, we generalize the $\mathsf{BSAD}$ algorithm to discounted MDPs with the traditional state-action rewards $r(s,a)\in[0,1]$. We first introduce the preliminaries of discounted MDPs:

\textbf{Discounted MDP:} The discounted MDP is represented by a tuple $\mathcal{M}=(\mathcal{S}, \mathcal{A}, \gamma, \mathbb{P}, \mu_0)$, where every step shares the same transition kernel $\mathbb{P}:\mathcal{S}\times\mathcal{A}\rightarrow\mathcal{S}$. Here $\gamma$ is the discount factor and there is no restart during the entire process. A policy $\pi:\mathcal{S}\rightarrow \mathcal{A}$ is defined as a mapping from states to actions. At each step $k$, the agent picks a policy $\pi$, observes the current state $s_k$, takes an action $a_k$ according to policy $\pi(s_k)$, the state then moves to the next state $s_{k+1}$ following transition kernel $\prob(\cdot|s_k, a_k)$. With traditional state-action rewards, we define the value function and Q-function:
\begin{align*}
    V^{\pi}(s) =& \expt\left[ \left.\sum_{i=1}^\infty \gamma^{i-1}r(s_i, \pi(a_i)) \right| s_1 = s\right],\\
    Q^{\pi}(s,a) =& r(s,a) + \expt\left[ \left.\sum_{i=2}^\infty \gamma^{i-1}r(s_i, \pi(a_i)) \right| s_1 = s, a_1 = a\right]
\end{align*}
Also we let $V^*$ and $Q^*$ denote the value functions of the optimal policy $\pi^*$ which maximizes the value function over the initial distribution. We also define the value function gap $\Delta(s,a) \equiv V^*(s) - Q^*(s,a)$. At the end of each step, the agent has the opportunity to query the human feedback using two sets of trajectories $\mathcal{D}_0$ and $\mathcal{D}_1$ with arbitrary cardinality $M$, and with arbitrary length of the trajectory, say $H$. Let $\mathcal{D}_0 = \{\tau^1_0, \tau^2_0, \cdots, \tau^{M_0}_0\}$, and $\mathcal{D}_1 = \{\tau^1_1, \tau^2_1, \cdots, \tau^{M_1}_1\}$, where each trajectory is a set of state-action pairs in sequence. The oracle will give feedback pointing to the dataset with a larger average reward similar to the episodic setting, i.e.,
\begin{align}
    \sigma = \mathsf{HumanFeedback}(\mathcal{D}_0, \mathcal{D}_1) = \argmax_{i \in \{0,1\}} \frac{1}{M}\sum_{m=1}^{M} \sum_{h=1}^H r(s_{h,i}^m, a_{h,i}^m),
\end{align}
where $s_{h,i}^m$ ($a_{h,i}^m$) is the $h$-th state (action) trajectory $m$ of dataset $\mathcal{D}_i$. The objective of discounted MDP is to identify the optimal policy $\pi^*$ without the reward feedback but using the human oracle.

\subsection{$\mathsf{BSAD}$ for Discounted MDPs}
\begin{algorithm}[t]
\caption{BASD for Discounted MDPs}\label{alg:BSAD_discounted}
\KwIn{planning horizon $H$}
    divide the time horizon into frames of length $H$, and treat each frame as an episode\;
    run BASD for episodic MDPs with planning horizon $H$ upon termination and obtain policy $\hat{\pi}$\;
    \Return $\hat{\pi}_1$.
\end{algorithm}

The method of generalizing $\mathsf{BSAD}$ for Episodic MDPs to a discounted setting is to segment the time horizon into frames with planing horizon $H$, as shown in Alg.~\ref{alg:BSAD_discounted}. Then, we treat the discounted MDP as episodic MDP, i.e., we treat the state $s_{H+1}$ as the initial state $s_1$ for the next episode. Then, we run the $\mathsf{BSAD}$ algorithm to obtain a candidate optimal policy $\{\hat{\pi}_h(s)\}_{h=1}^H$. When $H$ is chosen large enough, we show that the policy $\hat{\pi}_1$ is $\varepsilon$ close to the optimal policy $\pi^*$, i.e., $\expt_{\mu_0}[V^{\pi_1}(s)]\geq \expt_{\mu_0}[V^*(s) ]- \varepsilon$. Specifically, we chose the planning horizon to be:
\begin{align*}
    H = \Theta\left( \frac{\log\frac{1}{\varepsilon (1-\gamma)^2} }{1-\gamma} \right).
\end{align*}
Assume the Condorcet winner exists for all state $s$ and trajectory length $h\leq H$. We let $\overline{\Delta}_h^M(s,a)$ to be the probability gap for action $a$ and trajectories of length $H-h$ compared to the Condorcet winner of that state $s$, and $p^\pi_h(s|s')$ is the visitation probability of $s$ after $h$ steps starting from state $s'$ with policy $\pi$. Both definitions are analog to the definitions in episodic MDPs.

\begin{theorem}\label{thm:discounted}
    Fix $\delta$ and suppose $M$ is chosen large enough such that the Condorcet winner exists for all state $s$ and trajectory length $h\leq H$. Then with probability $1-\mathcal{O}(\delta)$, $\mathsf{BSAD}$ terminates within $K$ episodes and returns an $\varepsilon$-optimal policy with:
    \begin{align*}
        K = \tilde{\mathcal{O}}\left( \frac{SA^3M \log(\frac{1}{\delta})\log^3(\frac{1}{\varepsilon})}{(1-\gamma)^3\min_{h,s,a} \overline{\Delta}_h^M(s,a) ^2 \max_\pi \min_{s'} p^\pi_h(s|s')^2} \right).
    \end{align*}
\end{theorem}

\textbf{Proof of Correctness:} The sample complexity bound can be obtained from Theorem.~\ref{thm:sc_episodic} by substituting $H$ with our chosen value. Notice that for each manually constructed episode, the initial distribution is not $\mu_0$ anymore and depends on the previous episode. Therefore, we need to additionally replace the state visitation probability $p_h^\pi(s)$ in Theorem.~\ref{thm:sc_episodic} with the visitation probability of adversarial initial state, i.e., $\min_{s'}p^{\pi}_h(s|s')$. The rest is to prove the output policy $\hat{\pi}_1$ is $\varepsilon$-close to optimal, which we couple the infinite horizon MDP to a finite episodic horizon MDP with horizon $H$. By Theorem~\ref{thm:sc_episodic}, we know that $\hat{\pi}_1$ is with high probability the optimal policy for the first step in an $H$-step episodic MDP. Then, the problem reduces to bound the difference in the value function of a $H$-step episodic MDP and an infinite horizon MDP. The complete proof is in the appendix.

\subsection{Proof of Theorem.~\ref{thm:discounted}}
As mentioned in the main body, the sample complexity bound can be obtained through Theorem.~\ref{thm:sc_episodic} by substituting the parameters of discounted MDPs, specifically with $H = \tilde{\Theta}( \frac{1 }{1-\gamma} )$. Then, we prove that the output policy $\tilde{\pi}_1$ is with high probability $\varepsilon$-optimal. We couple the infinite horizon MDP $\mathcal{M}$ with a finite horizon MDP $\mathcal{M}^H$ with discount factor $\gamma$ and planning horizon $H$. Specifically, the transition kernel of $\mathcal{M}^H$ follows $\prob$ of the infinite horizon MDP, and the reward of each step is equal to the discounted reward of $\mathcal{M}$, i.e., $r_h^H(s,a) = \gamma^{h-1}r(s,a)$. Then, we use $V^{*,H}_h(s)$ and $Q^{*,H}_h(s,a)$ to denote the value function and Q function for the finite horizon MDP $\mathcal{M}^H$, and the optimal policy for it is denoted as $\pi^{*,H}$. First, from Theorem.~\ref{thm:sc_episodic}, with probability $1-\mathcal{O}(\delta)$, we have $\hat{\pi} = \pi^{*,H}$, then we show that $\pi_1^{*,H}$ is $\varepsilon$-optimal in the discounted MDP. Notice that $\pi^{*,H}$ is obtained by $\argmax_a Q^{*,H}_1(s,a)$, we have by the main Theorem of~\citep{singh1994valuefunctionbound}:
\begin{align*}
    \max_s \left| V^*(s) - V^{\pi^{*,H}_1}(s) \right| \leq \frac{2 \max_{s,a} \left| Q^{*,H}_1(s,a) - Q^*(s,a) \right| }{1-\gamma}.
\end{align*}
Notice that we have $\pi^{*,H}_1$ being the optimal policy of the $\mathcal{M}^H$, which implies for any $(s,a)$:
\begin{align*}
    Q_1^{*,H}(s,a) \geq& Q^{\pi^*,H}_1(s,a) \\
    = & r(s,a) + \expt\left[ \left.\sum_{i=2}^H \gamma^{i-1}r(s_i, \pi^*(a_i)) \right| s_1 = s, a_1 = a\right]\\
    \geq & r(s,a) + \expt\left[ \left.\sum_{i=2}^\infty \gamma^{i-1}r(s_i, \pi^*(a_i)) \right| s_1 = s, a_1 = a\right] - \frac{\gamma^H}{1-\gamma}\\
    =& Q^*(s,a) - \frac{\gamma^H}{1-\gamma}.
\end{align*}
Moreover, one can view the $Q^*(s,a)$ as the outcome of value iteration with infinite many steps, and view $Q_1^{*,H}$ as the outcome of value iteration with $H$ steps. Since the reward is non-negative, we have for any $(s,a)$ that $Q^*(s,a)$ dominates the optimal value function $Q^{*,H}_1(s,a)$ for any finite $H$:
\begin{align*}
    Q^*(s,a) - \frac{\gamma^H}{1-\gamma} \leq Q_1^{*,H}(s,a) \leq Q^*(s,a).
\end{align*}
We can substitute the difference of the Q function with the difference of the value function to obtain:
\begin{align*}
    \max_s \left| V^*(s) - V^{\pi^{*,H}_1}(s) \right| \leq \frac{2\gamma^H}{(1-\gamma)^2}.
\end{align*}
Let $H = \Theta( \frac{\log{\varepsilon^{-1} (1-\gamma)^{-2}} }{1-\gamma} )$,
we can obtain $\expt_{\mu_0}[V^{\pi^{*,H}_1}(s)]\geq \expt_{\mu_0}[V^*(s)] - \varepsilon$, and thus with probability $1-\mathcal{O}(\delta)$, policy $\hat{\pi}_1$ is $\varepsilon$-optimal.

\end{document}